\documentclass{article}

\usepackage[dvipsnames]{xcolor}
\usepackage{microtype}
\usepackage{graphicx}
\usepackage{subfigure}
\usepackage{booktabs} 

\usepackage{hyperref}

\usepackage[accepted]{icml2024}

\usepackage{amsmath}
\usepackage{amssymb}
\usepackage{mathtools}
\usepackage{amsthm}

\usepackage[capitalize,noabbrev]{cleveref}

\theoremstyle{plain}

\theoremstyle{definition}

\theoremstyle{remark}

\usepackage[textsize=tiny]{todonotes}

\usepackage{amsmath}
\usepackage{amsthm}
\usepackage{amssymb}
\usepackage{pifont}
\newcommand{\cmark}{\ding{51}}%
\newcommand{\xmark}{\ding{55}}%

\usepackage[utf8]{inputenc} 
\usepackage[T1]{fontenc}    
\usepackage{hyperref}       
\usepackage{url}            
\usepackage{booktabs}       
\usepackage{amsfonts}       
\usepackage{nicefrac}       
\usepackage{microtype}      

\usepackage{graphicx}
\usepackage{tabularx}
\usepackage{mdframed}
\usepackage{lipsum,multicol}
\usepackage{caption}

\usepackage{amsmath}
\usepackage{amsfonts}       
\usepackage{MnSymbol}
\usepackage{mathrsfs}
\usepackage{graphicx}
\usepackage{caption}
\usepackage{multirow}

\usepackage{cleveref}

\usepackage{multibib}
\newcites{app}{Appendix References} 

\usepackage{float}
\floatstyle{plain}
\newfloat{example}{thp}{lop}
\floatname{example}{Example}

\newcommand{\eq}{\begin{equation*}}
\newcommand{\en}{\end{equation*}}
\newcommand{\eqa}{\begin{eqnarray*}}
	\newcommand{\ena}{\end{eqnarray*}}
\newcommand{\eqn}{\begin{equation}}
\newcommand{\enn}{\end{equation}}
\newcommand{\be}{\begin{equation}}
\newcommand{\ee}{\end{equation}}
\newcommand{\eqan}{\begin{eqnarray}}
\newcommand{\enan}{\end{eqnarray}}

\newcommand{\nn}{\nonumber}

\newcommand{\s}{ {\bf s} }

\newcommand{\bfa}{ {\bf a} }
\newcommand{\bb}{ {\bf b} }

\newcommand{\y}{ {\bf y} }
\newcommand{\x}{ {\bf x} }
\newcommand{\z}{ {\bf z} }

\newcommand{\pmat}{\begin{pmatrix}}
\newcommand{\pman}{\end{pmatrix}}

\icmltitlerunning{Amortized Probabilistic Detection
of Communities in Graphs}

\begin{document}

\twocolumn[
\icmltitle{Amortized Probabilistic Detection
of Communities in Graphs}



\icmlsetsymbol{equal}{*}

\begin{icmlauthorlist}
\icmlauthor{Yueqi Wang}{equal,goog}
\icmlauthor{Yoonho Lee}{equal,stan}
\icmlauthor{Pallab Basu}{ww}
\icmlauthor{Juho Lee}{kaist}
\icmlauthor{Yee Whye Teh}{ox,dm}
\icmlauthor{Liam Paninski}{col}
\icmlauthor{Ari Pakman}{BGU}
\end{icmlauthorlist}

\icmlaffiliation{goog}{Google}
\icmlaffiliation{stan}{Stanford University}
\icmlaffiliation{col}{Culumbia University}
\icmlaffiliation{kaist}{KAIST}
\icmlaffiliation{ox}{Oxford University}
\icmlaffiliation{dm}{DeepMind}
\icmlaffiliation{ww}{University of the Witwatersrand}
\icmlaffiliation{BGU}{Ben-Gurion University of the Negev}

\icmlcorrespondingauthor{Ari Pakman}{pakman@bgu.ac.il}

\icmlkeywords{Machine Learning, ICML}

\vskip 0.3in
]



\printAffiliationsAndNotice{\icmlEqualContribution} 

\begin{abstract}
Learning community structures in graphs has broad applications across scientific domains.
While graph neural networks (GNNs) have been successful in encoding graph structures, 
existing GNN-based methods for community detection are limited by requiring knowledge of the number of communities in advance, in addition to lacking a proper probabilistic formulation to handle uncertainty. We propose a simple framework for amortized community detection, which addresses both of these issues by combining the expressive power of GNNs with recent methods for amortized clustering. Our models consist of a graph representation backbone that extracts structural information and an amortized clustering network that naturally handles variable numbers of clusters.
Both components combine into well-defined models of the 
posterior distribution of graph communities 
and are jointly optimized given labeled graphs. 
At inference time, the models yield parallel samples from 
the posterior of community labels, quantifying uncertainty in a principled way. 
We evaluate several models from our framework on synthetic and real datasets, and demonstrate improved performance compared to previous methods.
\end{abstract}

\section{Introduction}
\label{sec:introduction}

Community detection \cite{fortunato2010community,yang2013community} is a fundamental problem in network analysis with many applications such as finding communities in social graphs 
or functional modules in protein interaction graphs.
In machine learning, 
community detection is usually treated either as an unsupervised 
learning problem, or as posterior inference when a well-defined generative model
is assumed. 

Recent progress in graph neural networks (GNNs) has successfully extended the learning capabilities of deep neural networks to graph-structured data~\cite{bronstein2017geometric,hamilton_book,bronstein2021geometric}. 
Since community structures arise in real-world graphs, the classic problem of community detection can also benefit from the representation learning of GNNs, and this has been indeed the case for unsupervised approaches~\cite{wang2017community, cavallari2017learning,sun2019vgraph,jin2019community,tsitsulin2020graph}. 
However, neural models for community detection still face significant challenges, as recently reviewed in~\cite{liu2020deep,jin2021survey}. 
A limitation of existing models is the requirement of a fixed or maximum number of communities, typically encoded in the size of a softmax output.
This is a long-standing challenge in the field that constrains the model's ability to generalize to new datasets with a varying number of communities.
Moreover, the power of GNNs has not 
been applied to community detection as  
posterior inference, 
thus making it impossible to estimate the uncertainty on community assignment or the number of communities in networks, especially in noisy or ambiguous real-world data.

Training a neural network to learn posterior distributions 
is usually referred to as {\it amortized inference}~\cite{gershman2014amortized}. 
Concretely, denote a graph dataset as $\mathbf{x}$, the community labels of its $N$ nodes as~$c_{1:N}$, and assume the existence of a joint distribution~$p(\mathbf{x},c_{1:N})$. 
Amortization consists in training a 
neural network that takes as 
input a graph $\mathbf{x}$ 
and outputs a distribution over community assignments~$p(c_{1:N}|\mathbf{x})$.
In exchange for the initial cost of training a neural model, amortized inference offers several benefits compared to either
unsupervised approaches or other posterior inference methods.

Compared to unsupervised neural models for community discovery,
amortized approaches are more time-efficient
at test time, as they only require a forward pass evaluation on a pre-trained model, 
whereas unsupervised models typically go through an iterative optimization process for every test example~\cite{wang2017community, cavallari2017learning,sun2019vgraph,jin2019community,tsitsulin2020graph}. 
Moreover, the generic inductive biases encoded in unsupervised models might not be optimal for datasets with different structures, and amortized approaches can help by incorporating prior knowledge through model training.

\begin{figure}[t!]
\centering \includegraphics[width=.49\textwidth]{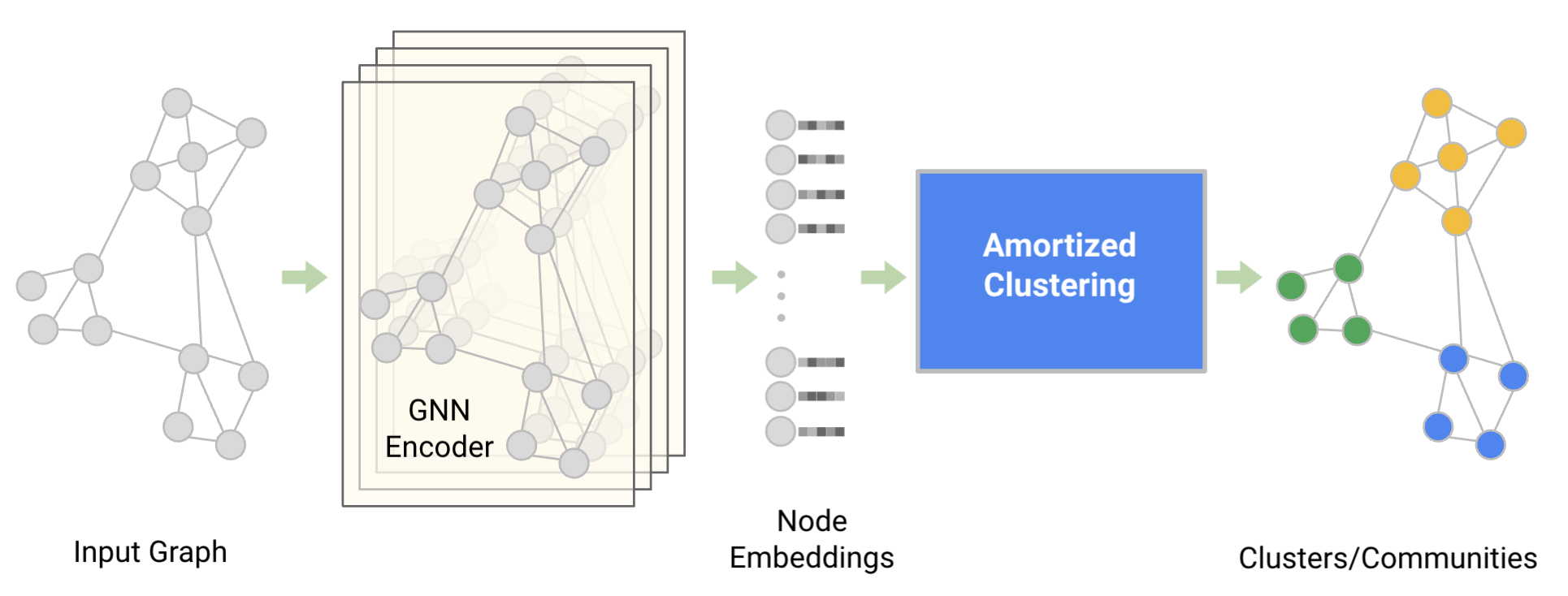}
 \caption{
 	{\bf Amortized Community Detection.} 
  }  
  \label{fig:amortized_community}
\end{figure}

Compared to other posterior inference methods,
the benefits of amortization are threefold.
First, well-trained models can produce i.i.d. posterior samples that match the accuracy of time-consuming Markov chain Monte Carlo (MCMC) methods at a fraction of the time~\cite{pakman2020}. 
Second, MCMC methods, as well as the faster but less accurate variational methods~\cite{blei2017variational}, usually require explicit forms for 
$p(c_{1:N})$ and $p(\mathbf{x}|c_{1:N})$ 
in order to sample or approximate the posterior $p(c_{1:N}|\mathbf{x})$, 
while amortization only requires {\it samples} from the joint generative model $p(\mathbf{x},c_{1:N})$,  
but not the putative model itself. 
Thus real-world labeled datasets can be 
easily incorporated into a probabilistic inference
setting without the need to fit a generative model. 
Finally, neural models that represent $p(c|\mathbf{x})$ yield explicit values for the 
probability of each sample $c$, a quantity 
usually unavailable in MCMC.
Thus when a single community labeling for a graph is required, one can simply pick the highest probability sample.


In this work we present an amortized 
framework for community discovery that combines GNNs with improved versions of amortized clustering architectures that naturally accommodate varying cluster numbers~\cite{DAC,pakman2020}, as illustrated in Figure~\ref{fig:amortized_community}. Our models are trained with labeled datasets containing varying numbers of communities, 
and yield samples from the posterior over community labels for test graphs of any size.


\section{Related Works}
\label{sec:related}

{\bf Neural models for community detection. } 
As mentioned above, almost all previous works 
on GNN-based community detection adopt an unsupervised learning approach.
The only previous supervised learning work addressing this task is~\cite{chen2018supervised}, 
which proposed two encodings for graph nodes specialized for community detection tasks (referred below as GNN and LGNN). 
For a fixed number $K$ of clusters, 
the models output for each node $i$ a softmax  $\phi_{i,c}$ for $c =1 \ldots K$.
The objective function is
\begin{align}  
I(\theta) = \min_{\pi \in {\cal S}_K} \sum_{i \in {\cal D}}  \log \phi_{i,\pi(c_i)} \,,
\label{eq:bruna_objective}
\end{align}  
where ${\cal D}$ indexes the dataset and ${\cal S}_K$ is the permutation group over the $K$ 
labels. Thus apart from fixing a maximum~$K$
in advance, the models
incur  the cost of evaluating $K!$ terms,
which makes them impractical for $K>8$ (see Figure~\ref{fig:training_time} below). 
More importantly, treating community discovery as
node classification ignores an important 
inductive bias of the problem,
as explained below in~\Cref{sec:acp}.

A related supervised learning problem was 
studied in~\cite{dwivedi2020benchmarking}
as a benchmark for GNN architectures.
But the task here is to find the members of each community {\it given a known initial node} for each community,
and is thus not directly comparable with 
our more difficult generic setting.

{\bf Amortized Clustering.} 
The amortized clustering models we adopt and extend in this work, reviewed in the next section, 
differ from previous works on supervised clustering~\cite{finley2005supervised,al2006adapting},
attention-based clustering~\cite{coward2020attention,ienco2020deep}
and neural network-based clustering (reviewed in~\cite{du2010clustering,aljalbout2018clustering,min2018survey}),
as these works focus on learning data features or similarity metrics as
inputs to traditional clustering algorithms,
while we instead model the posterior distribution of a generative model of clusters.

Supervised amortized probabilistic clustering was studied in~\cite{le2016inference,lee2018set,kalra2019learning}
for Gaussian mixtures with a fixed or bounded number of components. 
In those papers, the outputs of the network are the mixture parameters. 
This differs from the models we use in this work, which instead output the cluster labels of each data point,
and are not restricted to mixtures of Gaussians.


\begin{figure*}[t!]
\centering \includegraphics[width=.98\textwidth]
{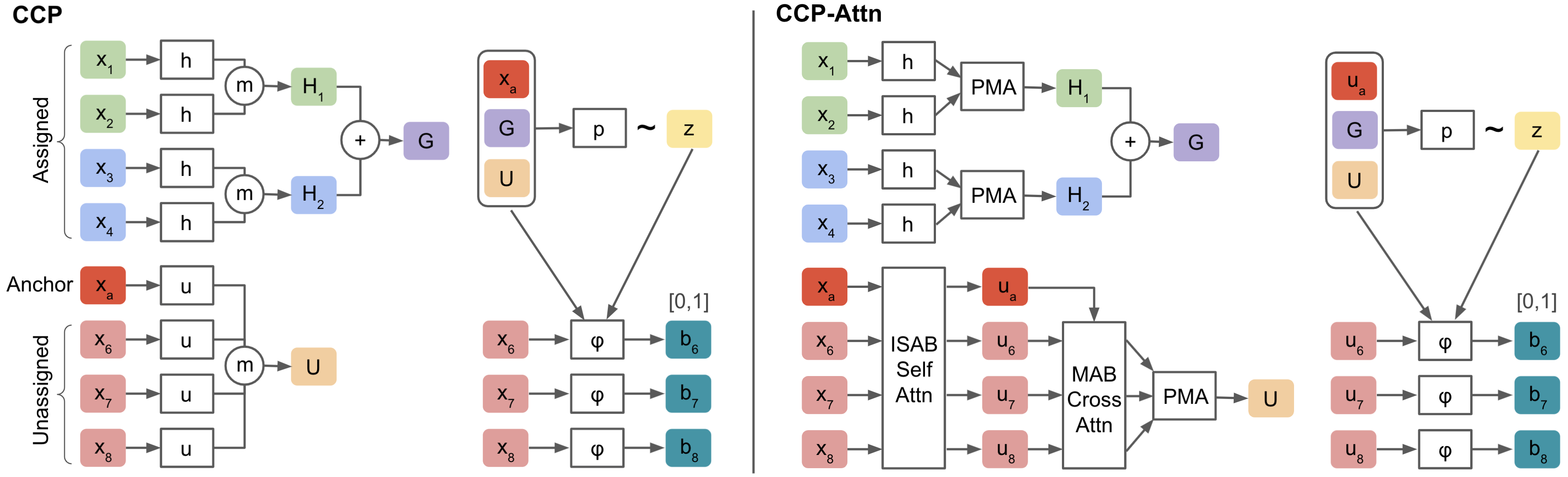}
 \caption{
  	{\bf CCP and CCP-Attn.}  {\it Left:} Architecture of the CCP model~\cite{pakman2020} for clusterwise amortized clustering. 
  	{\it Right:} Our proposed modification, CCP-Attn., where  
  	the mean aggregations \textcircled{\scriptsize m} used by CCP (see equation (\ref{eq:ccp_U_G})) are replaced  by Set Transformer attention modules from~\cite{lee2018set}. See Appendix \ref{app:ccp-attn-arch} for details. 
  }  
  \label{fig:ccp_model}
\end{figure*}

\section{Background}
\label{sec:background}

\subsection{Generative Models of Clusters}
\label{subsec:communities_as_clusters}

Our approach to community discovery is guided by its connection to generative models of clustering~\cite{mclachlan1988mixture}.
Let us denote the cluster labels as random variables~$c_i$, 
and assume a generative process
\begin{align} 
N &\sim p(N) 
\nn 
\\
\alpha_i &\sim p(\alpha_i), \quad i=1,2
\label{eq:cluster_prior}
\\
c_1 \ldots c_N &\sim p(c_{1},\ldots, c_{N}|\alpha_1) \,.
\nn 
\end{align}
The distribution in (\ref{eq:cluster_prior}) is
an exchangeable clustering prior and $\alpha_1, \alpha_2$ are hyperparameters.
We define the integer random variable~$K$ as the
number of distinct cluster indices~$c_i$. 
Note that  $K$  can take any value $K \leq N$, thus allowing for Bayesian nonparametric priors such as the Chinese Restaurant Process~(CRP) {(see~\cite{npbreview} for a review).}  This framework also encompasses cluster numbers $K<B$ for fixed $B$, such as Mixtures of Finite Mixtures~\cite{miller2018mixture}.

Given the cluster labels, standard mixture models assume 
$N$ observations $\mathbf{x} = \{x_i\}$ 
generated as
\begin{align} 
\begin{array}{rcl}
\mu_1 \ldots \mu_K & \sim &p(\mu_1, \ldots, \mu_K |\alpha_2) 
\label{eq:obs1} \\
x_i & \sim & p(x_i|\mu_{c_i}) \quad \text{for each} \quad i=1 \ldots N.
\end{array}
\end{align}
Here~$\mu_k$ controls the distribution of the $k$-th cluster,
e.g. as the mean and covariance of a Gaussian mixture component.


\subsection{Review of Amortized Clustering Models}
\label{sec:review_amortized_clustering}
Given $N$ observations $\mathbf{x} = \{x_i\}$, amortized clustering methods parameterize and learn a mapping from $\mathbf{x}$ to 
a distribution over the indices $\{c_i\}$. 
In this work we consider and compare three recent amortization models: 
\vskip 2mm
{\bf 1. Pointwise expansion (NCP Model)}.
Given $N$  data points  $\mathbf{x} = \{x_i\}$, 
we can sequentially expand the posterior distribution of labels as
\begin{align}  
p(c_{1:N}|\x) = p(c_{1}|\x)p(c_{2}|c_1, \x) \ldots p(c_{N}|c_{1:N-1}\x).
\label{eq:ncp_posterior}
\end{align}  
A neural architecture to model these factors, 
called the Neural Clustering Process (NCP), was proposed in~\cite{pakman2020}, 
and requires $O(N)$ forward evaluations for a full sample of cluster labels. Note that the range of values that $c_n$ can take is not fixed, 
since it depends on the previously assigned labels~$c_{1:n-1}$.
Therefore, the multinomial 
distribution $p(c_{n}|c_{1:n-1}\x)$ is not represented
using a standard softmax output, but a novel `variable-input softmax' 
introduced in~\cite{pakman2020}.

\vskip 2mm
{\bf 2. Clusterwise  expansion (CCP Model).}
An equivalent representation of the cluster labels~$c_{1:N}$ 
is given by the collection of $K$ sets $\s_{1:K}$, where each $\s_k$ contains the indices of points belonging to cluster $k$. For example, labels $c_{1:6}=(1,1,2,1,2,1)$ are equivalent to $\s_1=(1,2,4,6)$, $\s_2=(3,5)$.
Thus we can 
expand the posterior as 
\begin{align}  
p(\s_{1:K}|\x) = 
p(\s_1|\x) p(\s_2|\s_1,\x) \ldots p(\s_K|\s_{1:K-1}, \x).
\label{psk}
\end{align}  
with $p(c_{1:N}|\x) =  p(\s_{1:K}|\x)$. 
A neural model for these factors is the Clusterwise Clustering Process (CCP)~\cite{pakman2020} (Figure \ref{fig:ccp_model}).
To sample  from  $p(\s_k|\s_{1:k-1}, \x )$ we iterate two steps: (i) sample uniformly 
the first element~$x_a$  of 
$\s_k$  (called anchor) from the available points, (ii) 
choose which points join  $x_{a}$ 
to form $\s_k$ by sampling from 
\begin{align}  
p(\bb_{k}|x_a,\s_{1:k-1},\x) \,,
\label{eq:binary_posterior}
\end{align}  
where $\bb_{k}  \in \{0,1\}^{m_k}$ is 
a binary vector associated 
with the  $m_k$ remaining data points
$\{ x_{q_i}\}_{i=1}^{m_k}$. 
This  distribution 
depends both on these remaining points 
and on the assigned clusters $\s_{1:k-1}$
in permutation-symmetric ways, which 
can  be respectively captured by
symmetric encodings of the form~\cite{deep_sets}
\begin{align}  
U = \frac{1}{m_k} \sum_{i=1}^{m_k} u(x_{q_i}) \,,  \quad
G = \sum_{j=1}^{k-1} g\left(\frac{1}{|\s_j|} \sum_{i \in \s_j} h(x_i)\right) \,,
\label{eq:ccp_U_G}
\end{align}  
where $u, h,g$ are neural networks with vector outputs. 
Moreover, the binary distribution (\ref{eq:binary_posterior})
satisfies a form of conditional exchangeability~\cite{pakman2020} and
can be represented as 
\begin{align} 
&p(\bb_k| x_a,\s_{1:k-1},\x) \simeq 
\label{eq:marginal_z}
\\
&
\nn
\int \!\!
d\z_k  \!
\prod_{i=1}^{m_k} 
\varphi(b_{i}|\z_k, U, G, x_a, x_{q_i} )  
\, {\cal N}(\z_k|U, G, x_a) \,.
\end{align}  
Here $\varphi(b_i=1| \cdot)$ is a neural network 
with a sigmoid output, 
and 
the latent variable~$\z_k \in \mathbb{R}^{d_z}$
is a Gaussian, 
 with mean and variance
 parametrized by 
 neural networks with input~$(U, G, x_a)$. 
Since $b_{i}$'s in (\ref{eq:marginal_z})
are independent given  $\z_k$,
after sampling $\z_k$ all the $b_{i}$'s can be sampled in parallel.
Thus while a full sample of $\s_{1:K}$
 has cost $O(N)$, the heaviest
 computational burden, from network evaluations, 
scales as $O(K)$, 
as each
factor $p(\s_k|\s_{1:k-1}, \x)$ in~(\ref{psk}) needs $O(1)$ forward calls. Probability estimates for each sampled clustering configuration are provided in \cite{pakman2020}. 
The CCP architecture is illustrated in \Cref{fig:ccp_model}, left.  

\vskip 2mm
{\bf 3. Non-probabilistic clusterwise expansion (DAC Model).}
The Deep Amortized Clustering (DAC)~\cite{DAC} model 
is based on the expansion (\ref{psk}), 
but does not fully preserve the inductive biases
of the generative model. In its simplest version, 
it learns a binary classifier similar to~(\ref{eq:binary_posterior}), but assumes a form 
\begin{align}
    p\left(\mathbf{b}_k \mid \mathbf{x}, x_a\right) 
    \simeq \prod_{i=1}^{m_k} p\left(b_i \mid x_{i:m_k} , x_a\right) \,, 
\end{align}
i.e., it ignores previously sampled clusters $\s_{1:k-1}$
and the dependencies captured 
by $\z_k$  in (\ref{eq:marginal_z}). 

All  three  models are trained 
with labelled samples $(c_{1:N}, \x)$ 
by optimizing the model likelihood (for NCP and DAC)
or an evidence lower bound (ELBO) thereof (for CCP).
See~\cite{pakman2020, DAC} for details.

\begin{figure*}[t!]
\centering \fbox{
\includegraphics[width=0.15\linewidth]{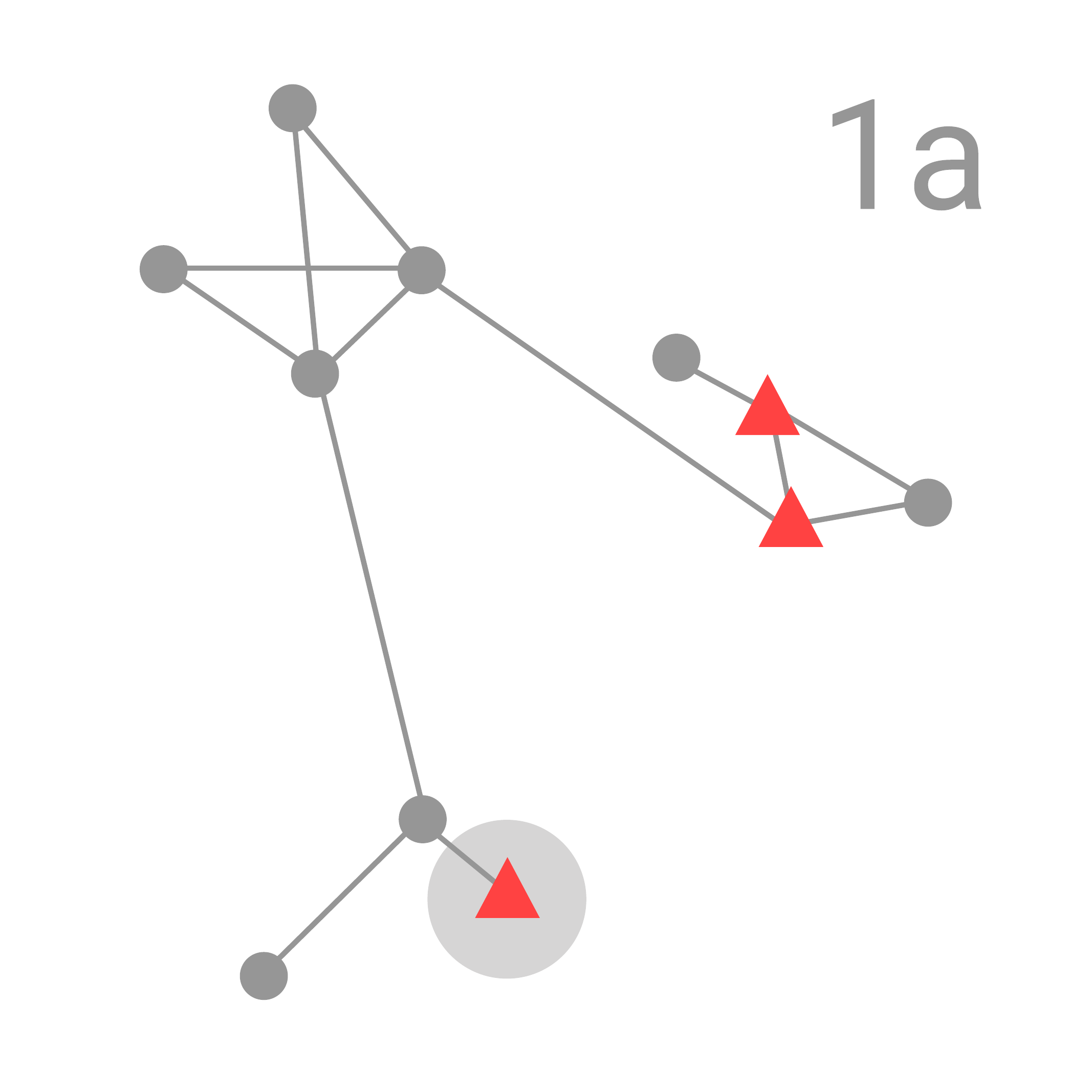}
\unskip\ 
\hfill \includegraphics[width=0.15\linewidth]{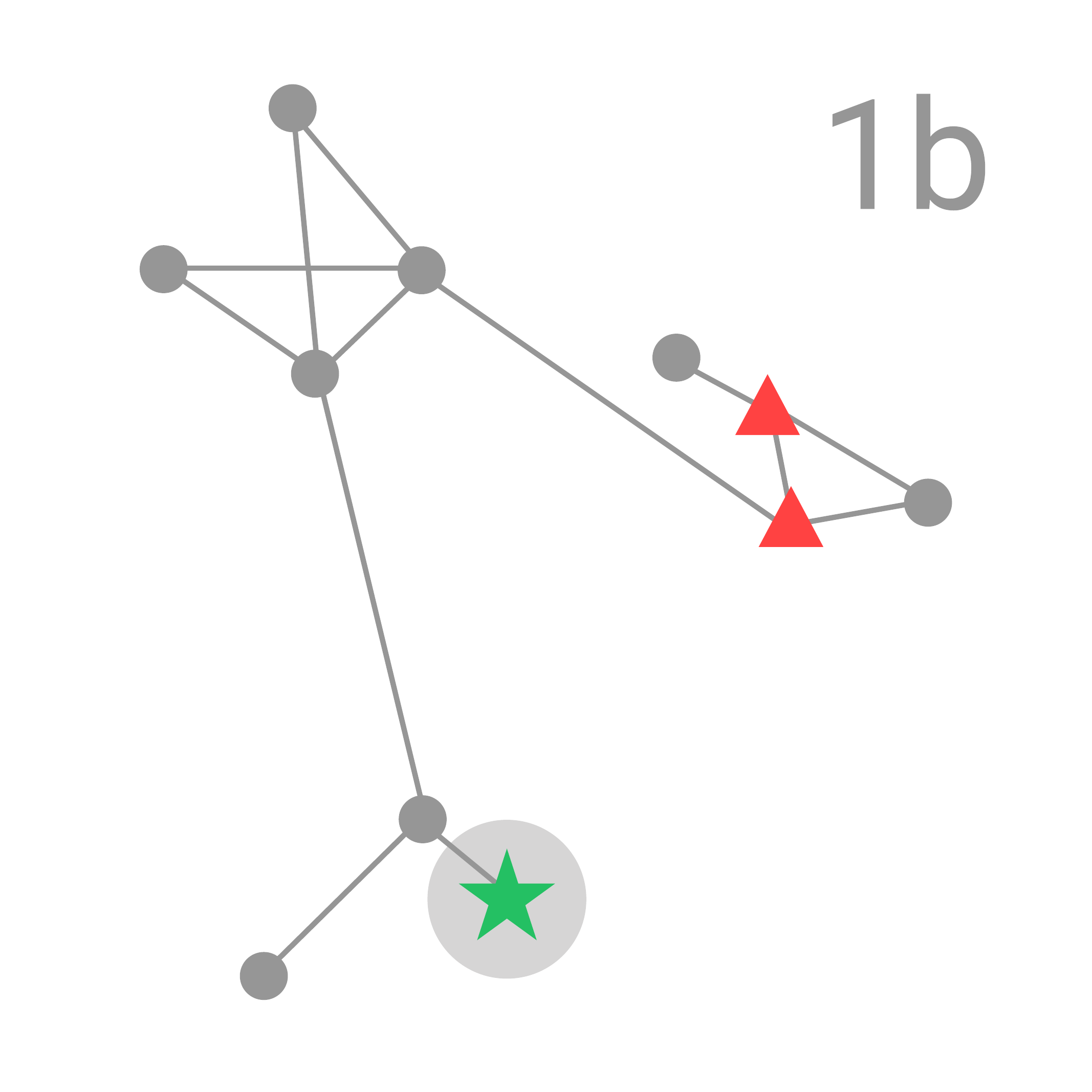}
\unskip\ \vrule
\hfill \includegraphics[width=0.15\linewidth]{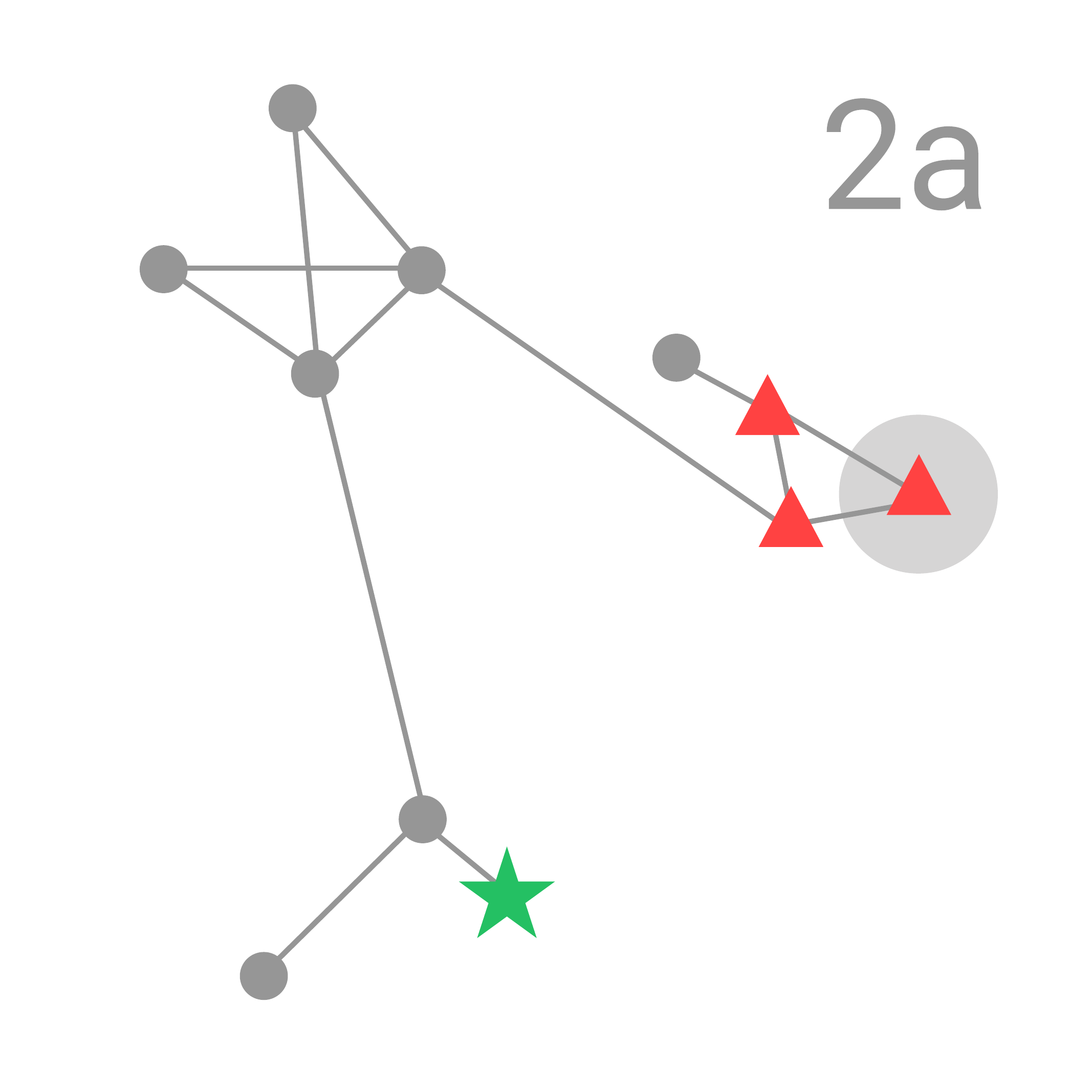}
\unskip\ 
\hfill \includegraphics[width=0.15\linewidth]{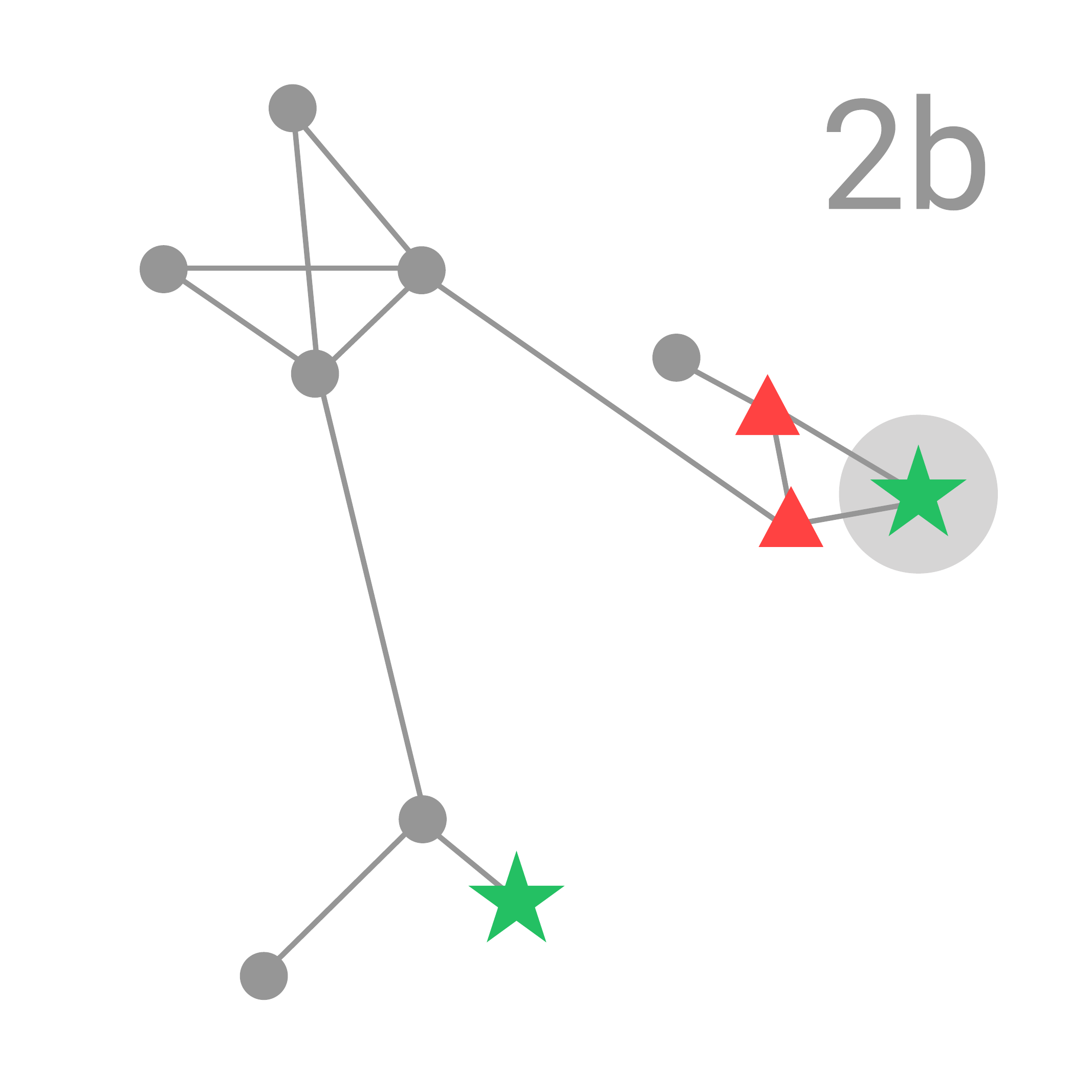}
\unskip\ 
\hfill \includegraphics[width=0.15\linewidth]{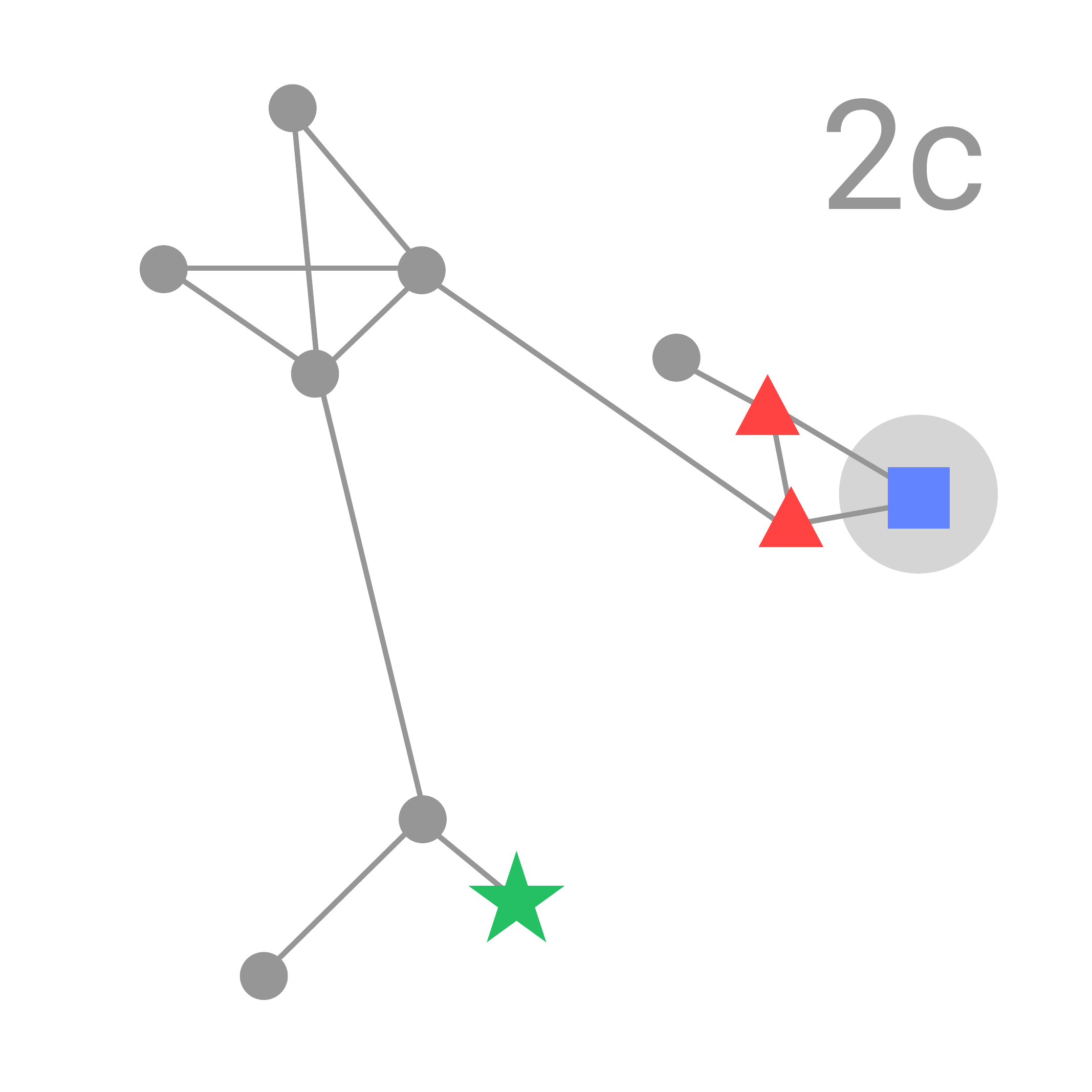}
}
\vspace{-2pt}
\caption{ {\bf Node-wise sampling (NCP Model).} 
{\it 1a-1b}: Two nodes have been already assigned to the same community ($c_1=c_2=1$, red triangles) and a randomly selected new node is sampled  from $p(c_3|c_{1:2},\x )$ to choose whether it joins them (1a, $c_3=1$), 
or creates a new community (1b, $c_3=2$, green star). {\it 2a-2c}: The previous node started its own community, 
and the next random node is sampled  from $p(c_4|c_{1:3},\x )$
to choose whether it joins any of the existing communities (2a, $c_4=1$;  2b, $c_4=2$) or creates a new community (2c, $c_4=3$, blue square). The procedure is repeated until all nodes have 
are sampled. 
} 
\label{fig:NCP}	
\end{figure*}

\subsection{Review of Graph Neural Networks}

The class of GNNs that we will consider are the widely used graph convolutional networks (GCNs), which
produce a feature vector $h_i^L \in \mathbb{R}^{d_h}$ associated to each node $i$ in a graph according to the formula
\begin{align}  
&
h_i^{\ell + 1} =
\sigma \left( W_1^{\ell}  h_i^{\ell} +
\sum_{j \in {\cal N}_i } \eta^{\ell}_{ij} W_2^{\ell} 
h_j^{\ell} \right), 
\label{eq:gcn_update}
\\
&
h_{i}^{\ell} \in \mathbb{R}^{d_h} \,,
\quad W_{1,2}^{\ell} \in \mathbb{R}^{d_h \times d_h} \,,
\quad \ell = 0\dots L-1 \,,
\nn 
\end{align}

where ${\cal N}_i$ is the set of neighbours of node $i$,
and $\eta_{ij}^{\ell} = f^{\ell}(h_{i}^{\ell}, h_{j}^{\ell})$ is a 
scalar or vector function (multiplied elementwise) 
that models the anisotropy of the encoding. 
Note that the update equation~(\ref{eq:gcn_update}) 
is independent of the graph size and is a
form of convolution since it shares the same weights across the graph. 
Each node requires also an initial input vector $h_i^{0}$, which can contain, e.g., node attributes. 
For recent book-length overviews of 
this vast topic see~\cite{hamilton_book, bronstein2021geometric}.

\section{Amortized Community Detection}
\label{sec:acp}
\subsection{Generative Model of Graphs with Communities}
For communities in graphs, 
we assume community labels for each node are generated as in the clustering prior (\ref{eq:cluster_prior}), 
followed by a generative model of edge data  ${\bf y} = \{ y_{ij} \}_{i,j=1}^N$~(e.g. direction or strength ), and possibly  node features~$\mathbf{f} = \{f_i\}_{i=1}^N$. 
For example, a popular generative model (without node features) is
\begin{align}
\label{eq:sbm1}
&
\phi_{k_1,k_2}\sim p(\phi | \beta)  \quad   k_1 \leq k_2 
\\
\label{eq:sbm2}
&
y_{ij}\sim \textrm{Bernoulli} (\phi_{c_i, c_j}) \,,   \,\,\,\, i\leq j\,,\quad i,j =1 \ldots N
\end{align} 
where $k_1,k_2=1 \ldots K$ and 
the  $y_{ij}$'s are binary variables
representing the presence or absence of an edge in 
the graph. 
Both the stochastic block model (SBM)~\cite{holland1983stochastic} and 
the single-type Infinite Relational Model~\cite{kemp2006learning,xu2006learning} 
use variants of the generative model~(\ref{eq:sbm1})-(\ref{eq:sbm2}). 

\subsection{Combining GNN with Amortized Clustering}

Our proposal in this work,
illustrated in \Cref{fig:amortized_community}, is to use a GNN to 
map observations ${\bf y}, {\mathbf{f}}$ to an embedding vector for each node, 
\begin{align}  
x_i = h_i^L({\mathbf{y}}, f_i) \,, \qquad i=1 \ldots N.
\end{align}  
We assume these node embeddings are approximate sufficient statistics for the posterior over labels,
\begin{align}  
p(c_{1:N}| {\bf y}, {\bf f} ) \simeq  p(c_{1:N}| x_{1:N}) \,,
\end{align}  
and use node embeddings $x_{1:N}$ as inputs to the amortized clustering modules reviewed in \Cref{sec:review_amortized_clustering}. 

For the initial node features $f_{1:N}$ as inputs to the GNN encoder, we choose the method of Laplacian eigenvector positional encoding~\cite{belkin2003laplacian,dwivedi2020benchmarking}, which takes the $m$ smallest non-trivial eigenvectors of the graph Laplacian matrix and encodes the graph positional information for each node.

Given labeled datasets of the form $(c_{1:N}, {\bf y}, {\bf f})$, both neural network modules (GNN and amortized clustering) are trained end-to-end by plugging in the objective function 
of the amortized model and the GNN-dependent inputs $x_i$.
Thus our proposed framework encourages the GNN to output node features that compactly represent the community structure, and the amortized clustering module uses such compact node features to identify communities.

\begin{figure*}
\centering \fbox{
\includegraphics[width=0.15\linewidth]{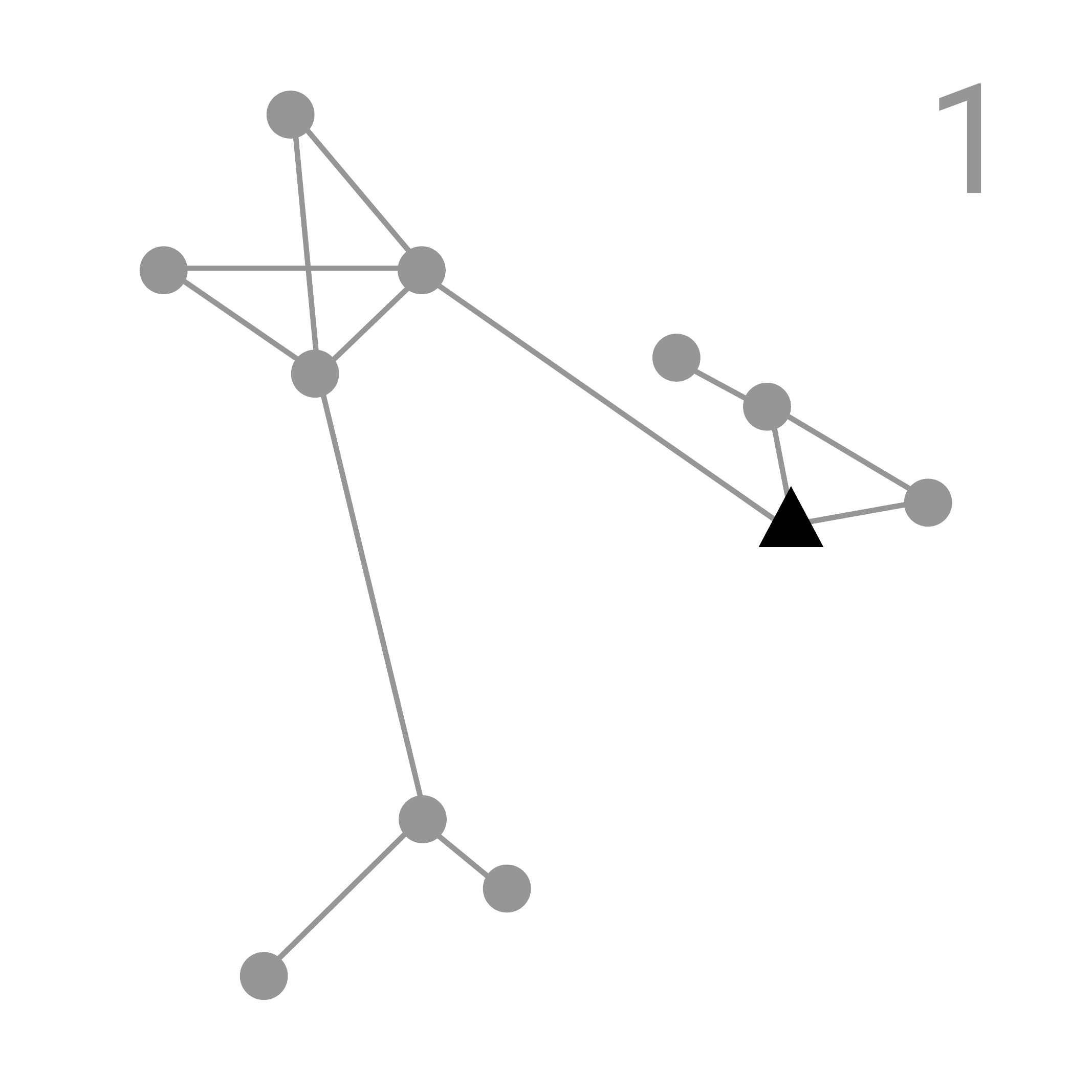}
\unskip\ \vrule
\hfill \includegraphics[width=0.15\linewidth]{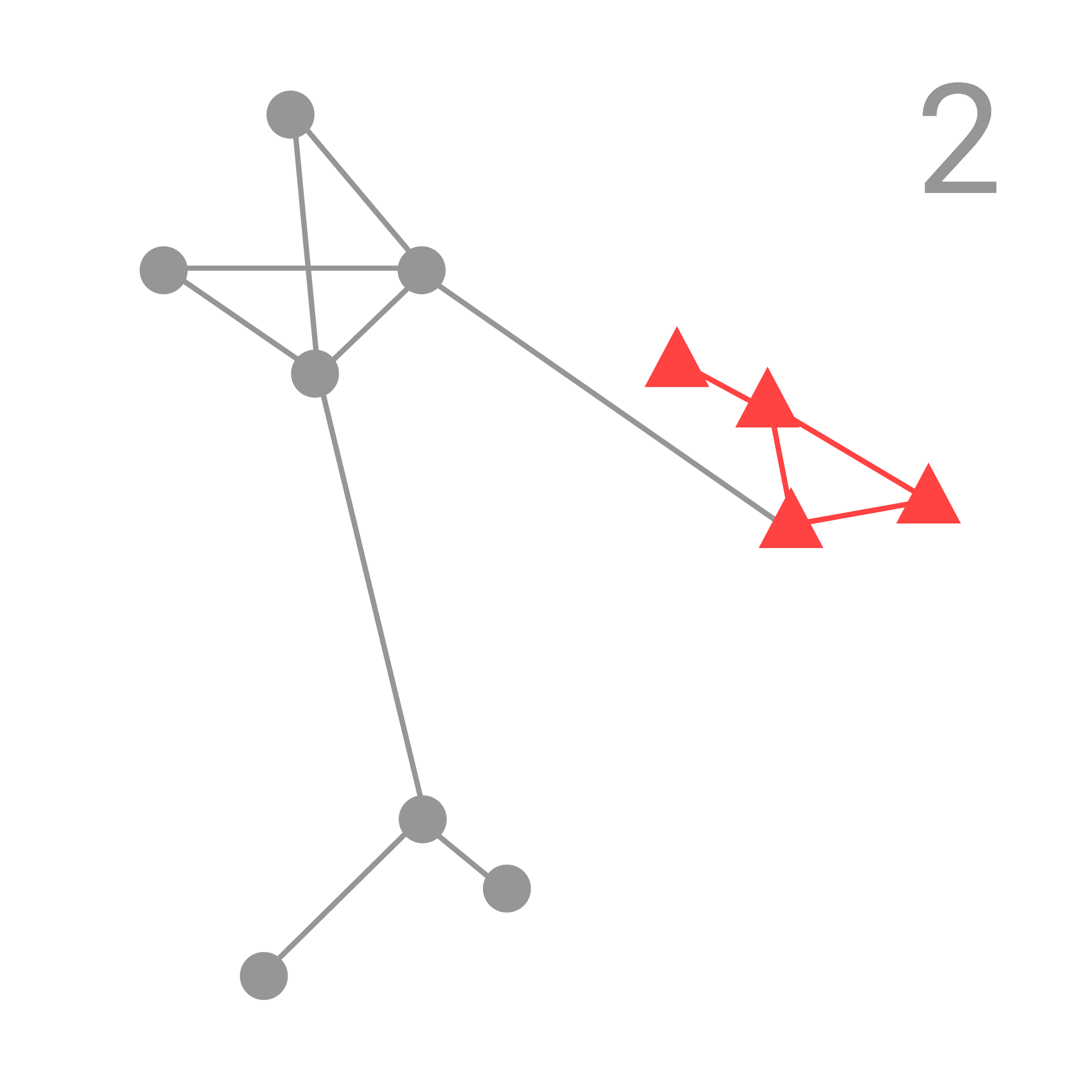}
\unskip\ \vrule
\hfill \includegraphics[width=0.15\linewidth]{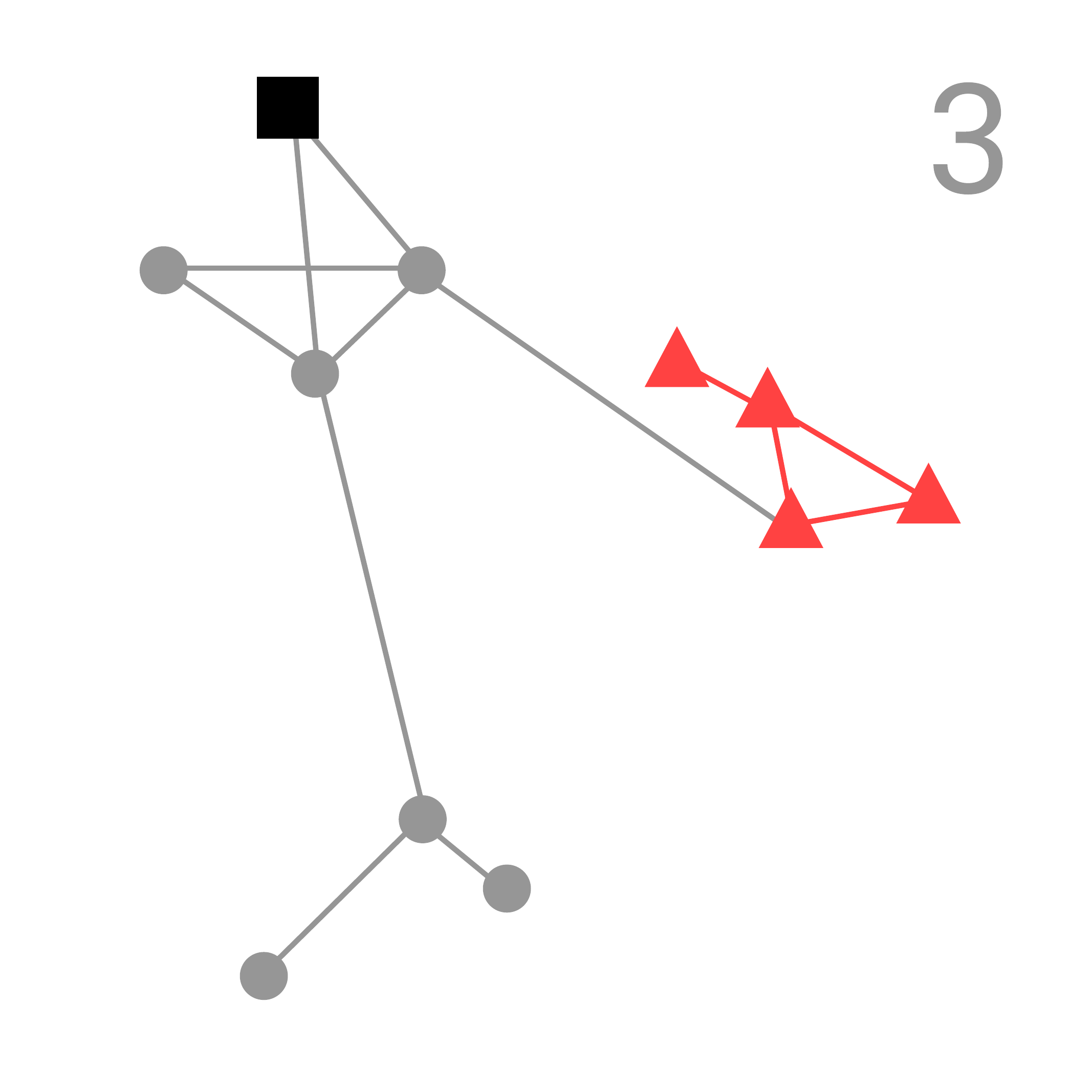}
\unskip\ \vrule
\hfill \includegraphics[width=0.15\linewidth]{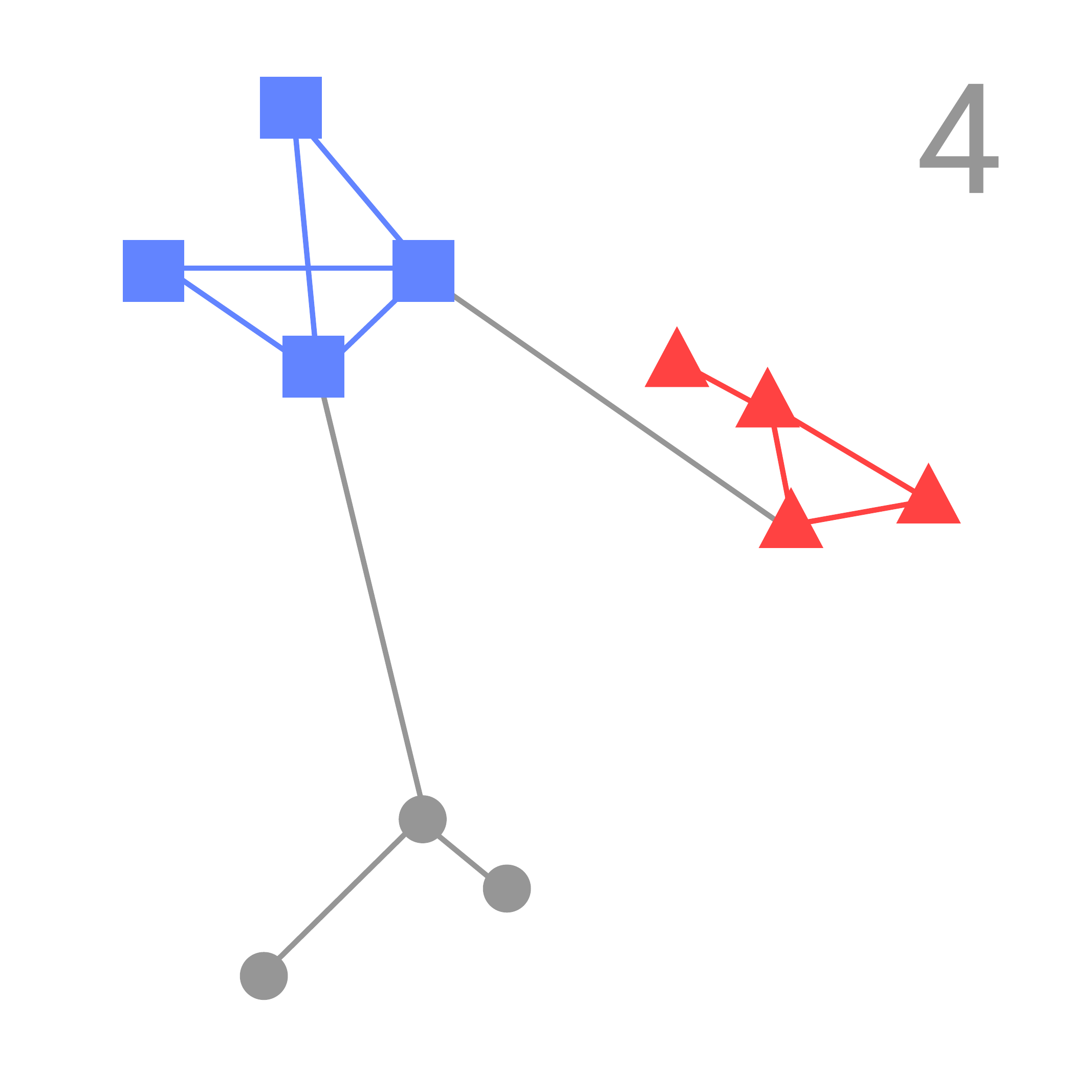}
\unskip\ \vrule
\hfill \includegraphics[width=0.15\linewidth]{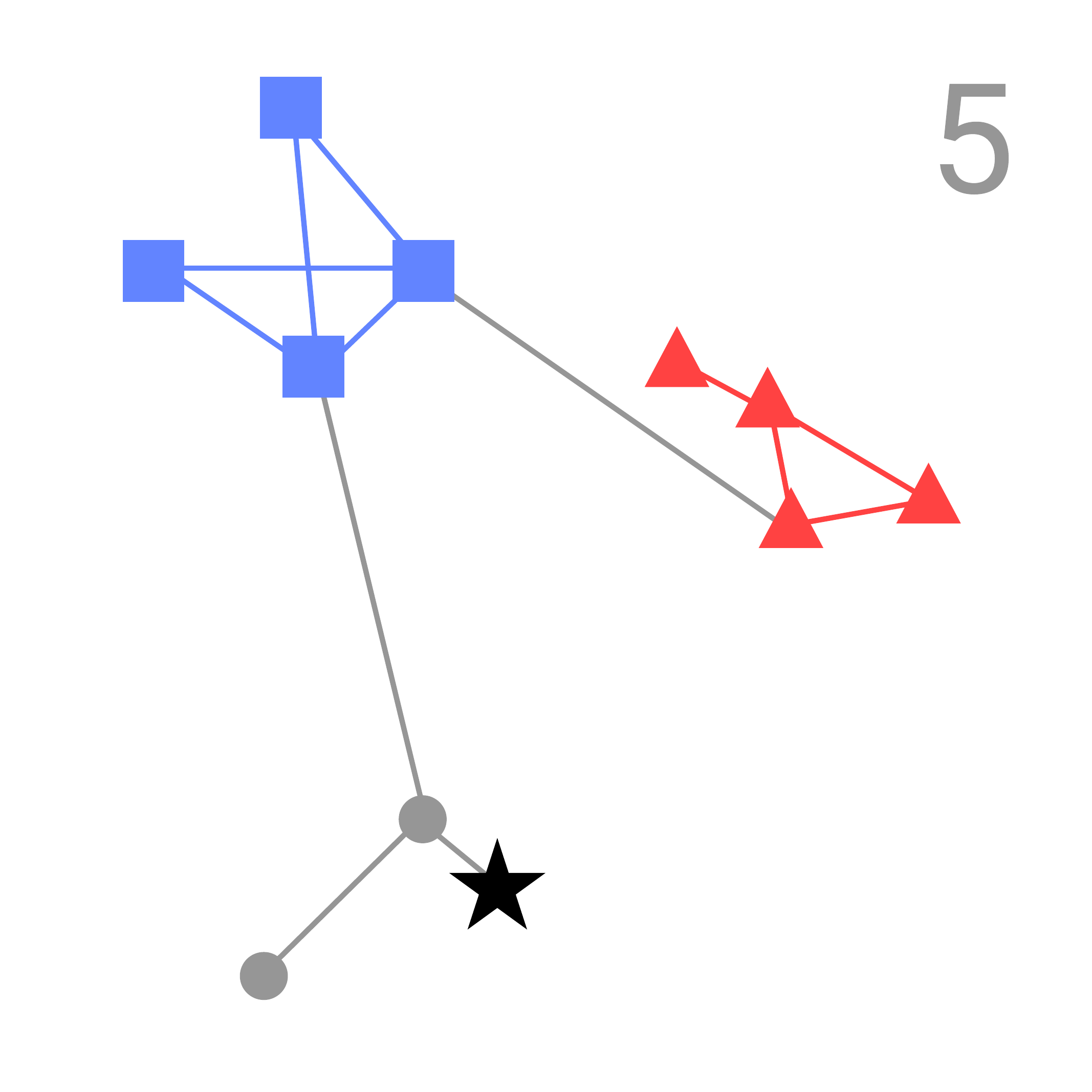}
\unskip\ \vrule
\hfill \includegraphics[width=0.15\linewidth]{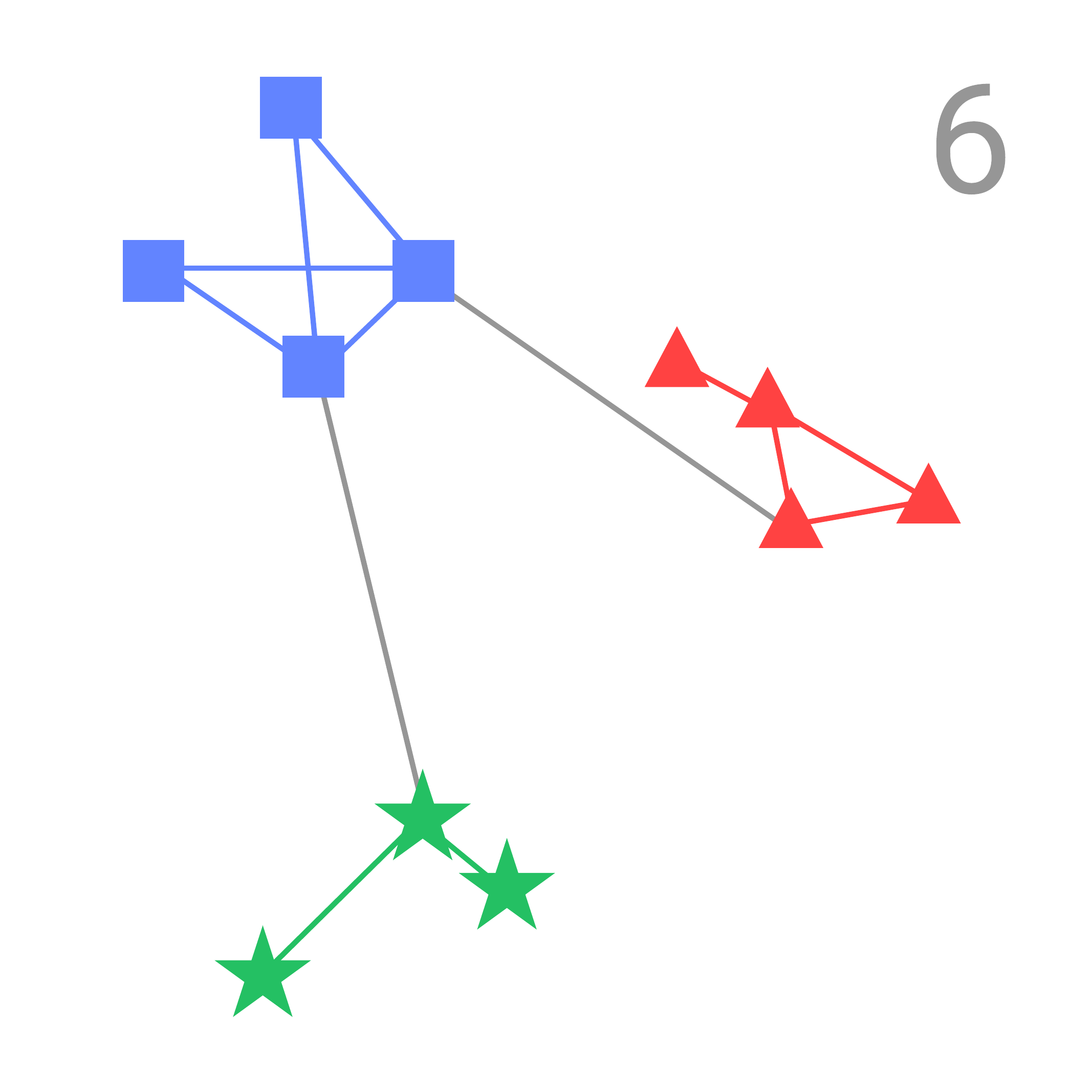}
}
\vspace{-2pt}
\caption{ {\bf Community-wise sampling (CCP and DAC Models).}  
{(1)}~The first element of community $\s_1$ (black triangle) is sampled uniformly,
and  the  available points  (grey dots)
are queried to join. 
(2) The first community~$\s_1$ is formed (red triangles).
{(3)}~The first element of $\s_2$ (black square)  is sampled uniformly from unassigned points.
{(4)}~The second community $\s_2$ is formed (blue squares).
(5)-(6) We repeat this procedure until no unassigned points are left. 
In CCP, the binary queries are correlated, 
but become independent conditioned on a latent vector, 
thus allowing parallel sampling (eq.(\ref{eq:marginal_z})).
} 
\label{fig:CCP}	
\end{figure*}

We visualize the progression of label assignments in a graph with community structure for the two different posterior expansions of the amortized clustering module.
\Cref{fig:NCP} and \Cref{fig:CCP} respectively illustrate 
the node-wise expansion used by NCP~(eqn.~\ref{eq:ncp_posterior}) and 
the community-wise expansion used by CCP and DAC~(eqn.~\ref{psk}).
We note that the simplifying assumptions made by DAC
about label independence are similar to those made 
by models that treat community discovery as 
classification (such as LGNN~\cite{chen2018supervised} or CLUSTER~\cite{dwivedi2020benchmarking}).

\subsection{Adding Attention to Amortized Clustering}
To increase the expressivity of the amortized clustering models,
we added attention modules to the NCP and CCP models by replacing mean and sum aggregation steps with variants of multi-head attention (MHA) blocks~\cite{vaswani2017attention}. For DAC~\cite{DAC}, 
the original formulation already has attention modules. More concretely, we used
three MHA-based modules with learnable parameters defined in~\cite{lee2018set}
that act on sets, named Multihead Attention Block (MAB), Pooling by Multihead Attention (PMA), and Induced Self-Attention Block (ISAB). 

We call the amortized models with attention NCP-Attn and CCP-Attn. 
\Cref{fig:ccp_model}, right, illustrates how CCP-Attn modifies CCP by replacing the mean operations in equation~(\ref{eq:ccp_U_G}) by attention modules. 
For NCP, attention-based aggregation is harder to incorporate due to the $O(N)$ forward evaluations. Instead, we replaced the $h$ and $u$ functions in (\ref{eq:ccp_U_G}) with ISAB self-attention layers across all input points, so that it is evaluated only once before the $O(N)$ iterations.
We describe the details of the attention modules and their use in the NCP and CCP architectures in Appendix \ref{app:attention}.

\newcommand{\mycheck}{\textcolor{OliveGreen}{\cmark}}
\newcommand{\myx}{\textcolor{OrangeRed}{\xmark}}
\newcommand{\shorttime}{\textcolor{OliveGreen}{$O(K)$}}

\begin{table}[h!] \centering
\caption{
{\bf Comparison community detection methods.}
The column `Pr.' indicates whether the method is probabilistic.
($*$): LGNN takes only one evaluation because it assumes fixed or maximum $K$ and is not directly comparable.
}
\label{tab:comparison}
\vspace{2pt}
\resizebox{1\linewidth}{!}{ 
\begin{tabular}{lccccc}
\toprule
Method & Learning & $K$ Constraints & Pr. & Cost \\
\midrule
DMoN \cite{tsitsulin2020graph} & Unsupervised & Max $K$ & \myx & N/A \\
LGNN/GNN \cite{chen2018supervised} & Supervised & $K < 8$ & \myx & $1^*$ \\
\midrule
GCN-DAC \cite{DAC} (ours) & Supervised & Varying $K$ & \myx & \shorttime \\
GCN-NCP \cite{pakman2020} (ours) & Supervised & Varying $K$ & \mycheck & \textcolor{OrangeRed}{$O(N)$} \\
GCN-CCP \cite{pakman2020} (ours) & Supervised & Varying $K$ & \mycheck & \shorttime \\
\bottomrule
\end{tabular}
}
\end{table}

\section{Experiments}
\label{sec:applications}
\renewrobustcmd{\bfseries}{\fontseries{b}\selectfont}
\newrobustcmd{\BF}{\bfseries}
\newcommand{\msd}[2]{{#1}{\tiny $\pm${#2}}}
\newcommand{\msdz}[2]{{#1}}

\begin{figure}[h!]
\centering
\includegraphics[width=\linewidth]{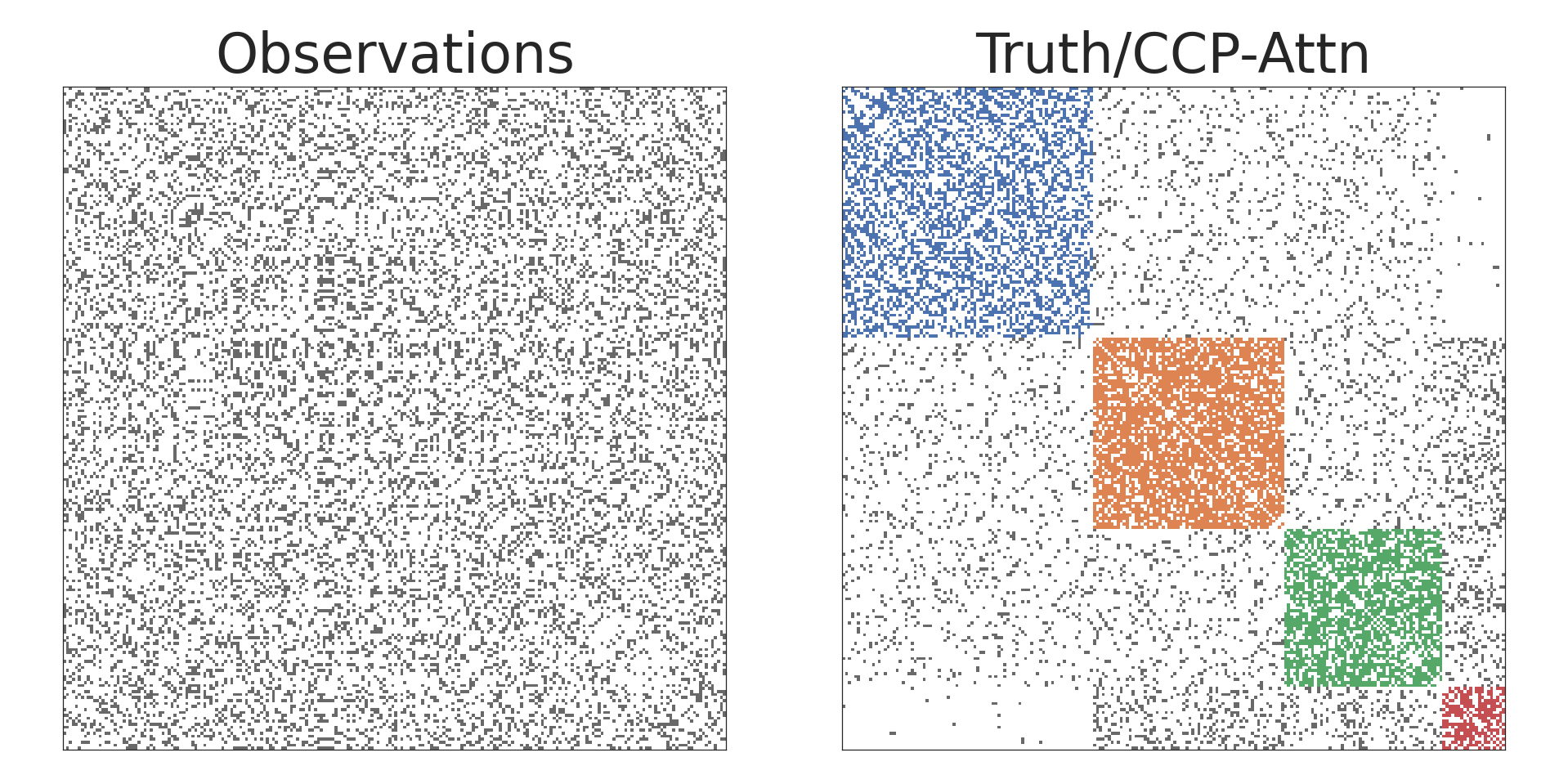}
\captionof{figure}{{\bf General SBM.} {\it Left:} Observations 
 	($N=222$). {\it Right:} Exact community recovery by CCP-Attn. }
\label{fig:sbm_beta_crp_example}
\end{figure}

\begin{table} 
\captionof{table}{{\bf Clustering SBM using CCP-Attn with different input embeddings and GCN encoders.}  The SBM data contains $1\sim16$ communities. 
Pos Enc: positional encoding; Rand Feat: random features.}
\footnotesize
\resizebox{\linewidth}{!}{ 
\begin{tabular}{lllcc}
\toprule%
            Input & GCN & AMI{\tiny $\times 100$} & ARI{\tiny $\times 100$}  & ELBO \\ 
\midrule%

Rand Feat. & GraphSAGE & \msd{57.0}{0.5} & \msd{54.4}{0.9} & \msd{-120}{3}
\\   
\midrule%
Rand Feat. & GatedGCN & \msd{66.5}{1.9} & \msd{65.5}{2.5} & \msd{-93}{2} 
\\ 

\midrule%
Pos Enc. & GraphSAGE &  \msd{82.8}{0.2} &  	\msd{82.1}{0.1}   & \msd{-45}{0.1}	 
\\      

\midrule%
Pos Enc. & GatedGCN & \BF \msd{89.6}{0.4}  & \BF \msd{88.5}{0.6} & \BF \msd{-23}{0.4} 
\\ 

\bottomrule
\end{tabular}
}
\label{tab:sbm_beta_crp_result}
\end{table}

We present several experiments to illustrate our framework.
Comparisons are made with LGNN/GNN~\cite{chen2018supervised}, a previous supervised
model for community detection, and DMoN~\cite{tsitsulin2020graph}, a recent
unsupervised neural model for community detection with state-of-the-art performance compared to other unsupervised approaches. 
\Cref{tab:comparison} compares all the models we considered.

In the examples below, 
the reported single estimates from our NCP and CCP-based models 
correspond to the maximum a posteriori (MAP)
sample maximizing the probability $p(c_{1:N}|\x)$,
estimated from multiple GPU-parallelized posterior samples. A default sampling size of 15 is used unless noted otherwise. We measure similarity between inferred and true community labels using the Adjusted Mutual Information~(AMI)~\cite{vinh2010information}
and/or Adjusted Rand Index~(ARI)~\cite{ARI} scores, which 
take values in $[0,1]$, with~1 corresponding to perfect matching. 
Other metrics are indicated in each case.

To study a generic form of community detection, in all our examples 
the data is given by undirected edge connectivity, and nodes
had no attributes. Details of the neural architectures and experiment setup appear in the Appendix.
Code to reproduce our experiments 
is available at \url{https://github.com/aripakman/amortized_community_detection}.

\subsection{Datasets}
\label{sec:datasets}

We consider two main types of datasets for evaluating our community detection models.

{\bf General SBM.}
The SBM generative model is eqs.(\ref{eq:sbm1})-(\ref{eq:sbm2}) and generated with $N \sim \textrm{Unif} [50,350]$, $c_1 \ldots c_N \sim  \textrm{CRP}(\alpha) $,   $p = \phi_{k_1,k_2} \sim$ \textrm{Beta}(6, 4) for $k_1 = k_2$, and $q = \phi_{k_1,k_2} \sim$ \textrm{Beta}(1, 7) for $k_1 \neq k_2$.
        Here CRP($\alpha$) is a  Chinese Restaurant Process with concentration  $\alpha = 3.0$. 
        The graphs contain varying ($K=1\sim16$) numbers of communities with 
        varied connection probabilities, as illustrated in Figure~\ref{fig:sbm_beta_crp_example}. The SBM data tests the models' ability to learn from a generative model.

{\bf Real-world SNAP datasets.} The SNAP datasets \cite{snapnets} contains real-world networks from social connections, collaborations, and consumer products. Unlike the SBM, we lack here an explicit generative model, but that is not a problem since all we need for training are ground-truth community labels.
We used two SNAP datasets (DBLP, Youtube) and followed the data preparation of \cite{yang2015defining} and \cite{chen2018supervised} to extract subgraphs composed of 2 to 4 non-overlapping communities.

\subsection{GNN Encoders and Input Node Features}

To find an effective GNN encoder, we considered two GCN variants:
(i) the isotropic GraphSAGE~\cite{hamilton2018inductive} (with $\eta_{ij}^{\ell}=1$ in (\ref{eq:gcn_update})) and (ii) the anisotropic GatedGCN~\cite{bresson2018residual}, shown to perform well on node classification benchmarks~\cite{dwivedi2020benchmarking}.
As input features of the nodes 
we used a 20-dimensional Laplacian eigenvector positional encoding \cite{belkin2003laplacian,dwivedi2020benchmarking}, and compared it with a baseline of 20-dimensional random vector sampled from $\mathcal{N}(0,1)$.
Using the CCP-Attn amortized clustering model and training on the SBM dataset, we show in Table~\ref{tab:sbm_beta_crp_result} that
GatedGCN and positional encoding consistently perform better than GraphSAGE and random input features, and we use them for other experiments below.

\begin{table*}[t] 
\caption{
{\bf Clustering performance on synthetic SBM and real-world SNAP datasets.} Each dataset contains either fixed or varying numbers of communities as denoted by $K$. For models that require fixing a maximum $K$, results for different $K$ values are shown.
The AMI and ARI scores are multiplied by 100. Times are in the unit of seconds. 
Means and standard deviations of metrics are from $3-6$ independently trained models (DMoN was run once on each dataset). The standard deviation of inference time is small and thus omitted.
OOM: out-of memory.
}
\label{tab:sbm_and_snap_results}
\centering
\resizebox{\linewidth}{!}{ 
\begin{tabular}{lc ccc cc cc ccc ccc}
\toprule
    & 
    & \multicolumn{3}{c}{SBM ($K=1\sim16$)}
    & \multicolumn{2}{c}{DBLP ($K=3$)}
    & \multicolumn{3}{c}{DBLP ($K=2\sim4$)}
    & \multicolumn{2}{c}{Youtube ($K=3$)}
    & \multicolumn{3}{c}{Youtube ($K=2\sim4$)} \\
    \cmidrule(lr{1em}){3-5}  
    \cmidrule(lr{1em}){6-7}  
    \cmidrule(lr{1em}){8-10}  
    \cmidrule(lr{1em}){11-12}  
    \cmidrule(lr{1em}){13-15}  
Model  
& Max $K$   
& AMI & ARI & Time 
& AMI & ARI 
& AMI & ARI & Time 
& AMI & ARI 
& AMI & ARI & Time 
\\

\midrule
DMoN \cite{tsitsulin2020graph} & 16
    & 60.0 & 46.7 & 12.1
    & 52.8 & 33.4 
    & 39.7 & 25.3 & 10.7
    & 40.6 & 18.2 
    & 32.4 & 14.8 & 11.4 \\

 & 4
    & - & - & -
    & 54.6 & 47.1
    & 45.1 & 36.5 & 10.0
    & 50.2 & 43.0
    & 40.2 & 29.3 & 10.1 \\

\midrule
LGNN \cite{chen2018supervised} & 7
    & OOM  & OOM & OOM
    & - & - 
    & \msd{54.2}{2.1} & \msd{51.2}{2.3} & 0.44
    & - & - 
    & \msd{68.7}{0.7} & \msd{71.3}{1.1} & 0.28
\\


& 4
    & - & - & -
    & - & - 
    & \msd{52.7}{2.2} & \msd{49.4}{2.3} & 0.42
    & - & - 
    & \msd{68.4}{1.4} & \msd{71.3}{1.1} & 0.27
\\
& 3
    & - & - & -
    & \msd{66.8}{1.3} & \msd{62.7}{1.6}
    & - & - & -
    & \msd{76.7}{0.8} & \msd{78.4}{0.8}
    & - & - & -
\\

GNN \cite{chen2018supervised} 
& 7
    & \msd{76.2}{0.9} & \msd{75.9}{1.2} &  0.15
    & - & - 
    & \msd{51.5}{2.5} & \msd{48.4}{2.5} & 0.035
    & - & - 
    & \msd{67.3}{3.8} & \msd{69.5}{4.0} & 0.033
\\


& 4
    & - & - & -
    & - & - 
    & \msd{55.4}{3.3} & \msd{52.2}{4.1} & 0.036
    & - & - 
    & \msd{67.5}{1.2} & \msd{69.7}{1.4} & 0.033
\\
& 3
    & - & - & -
    & \msd{66.3}{1.7} & \msd{62.7}{1.7}
    & - & - & -
    & \msd{76.2}{2.6} & \msd{78.0}{2.0}
    & - & - & -
\\

\midrule
GCN-NCP & -
    & \msd{89.6}{0.2} & \msd{88.3}{0.3} &  0.88
    & \msd{74.3}{2.1} & \msd{71.0}{2.1}
    & \msd{54.8}{0.3} & \msd{51.0}{0.3} & 0.54
    & \msd{84.5}{0.9} & \msd{84.6}{0.9}
    & \msd{64.7}{3.4} & \msd{64.2}{3.3} & 0.43
\\

GCN-NCP-Attn & -
    & \BF \msd{89.9}{0.4} & \BF \msd{88.9}{0.6}	 &  1.0
    & \msd{82.1}{0.6} & \msd{78.7}{0.5}
    & \msd{56.4}{2.1} & \msd{52.4}{2.0} & 0.47
    & \msd{84.3}{0.9} & \msd{84.5}{1.0} 
    & \msd{69.2}{4.1} & \msd{68.9}{4.3} & 0.43
\\

\midrule
GCN-DAC & -
    & \msd{87.1}{0.1} & \msd{87.1}{0.1} &  0.16
    & \msd{79.2}{1.4} & \msd{74.4}{1.8}
    & \msd{60.3}{0.7} & \msd{55.9}{0.8} & 0.073
    & \msd{82.6}{0.5} & \msd{82.6}{0.6}
    & \msd{73.7}{1.3} & \msd{73.5}{1.5} & 0.081
\\
GCN-CCP  & -
    & \msd{88.8}{0.5} & \msd{87.3}{0.7} &  0.3
    & \msd{82.1}{0.5} & \msd{78.6}{0.6}
    & \msd{62.4}{3.7} & \msd{57.6}{3.4} & 0.12
    & \msd{83.7}{0.8} & \msd{83.8}{0.7}
    & \msd{76.2}{1.9} & \msd{76.2}{2.1} & 0.13
\\
GCN-CCP-Attn  & -
    & \msd{89.6}{0.4} & \msd{88.5}{0.6} &  0.59
    & \BF \msd{89.2}{0.4} & \BF \msd{79.7}{0.5}
    & \BF \msd{65.3}{1.2} & \BF \msd{59.7}{1.8} & 0.22
    & \BF \msd{85.5}{0.3} & \BF \msd{85.7}{0.4}
    & \BF \msd{79.2}{0.4} & \BF \msd{79.3}{0.4} & 0.22
\\
\bottomrule \end{tabular} 
}
\end{table*}

\begin{figure}[b!]
\centering 
\includegraphics[width=.48\textwidth]{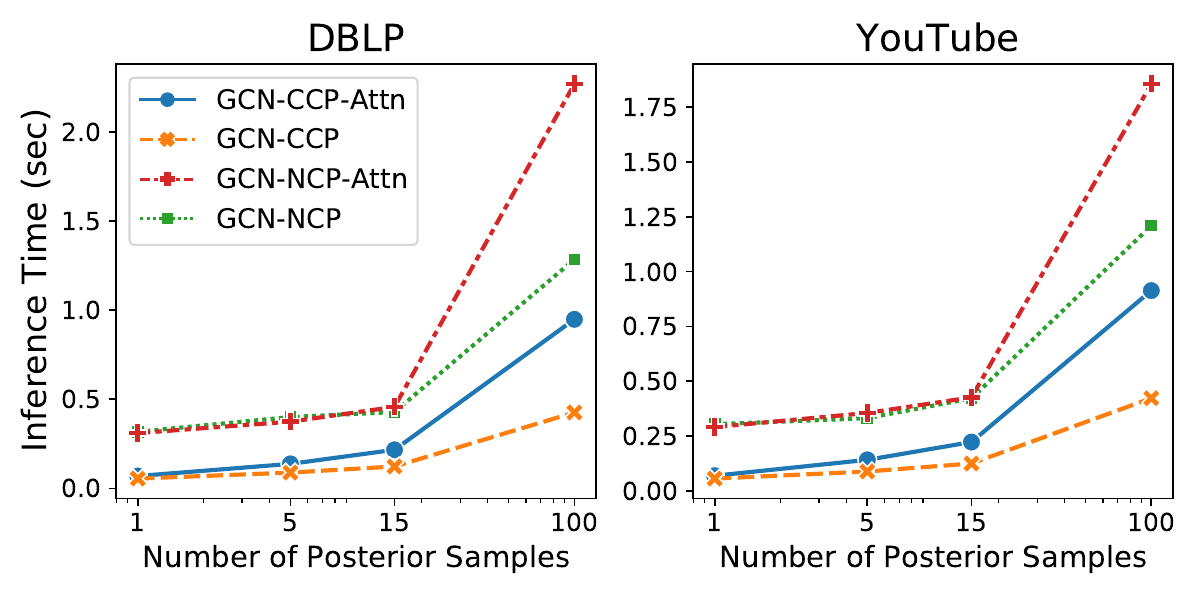}
\includegraphics[width=.48\textwidth]{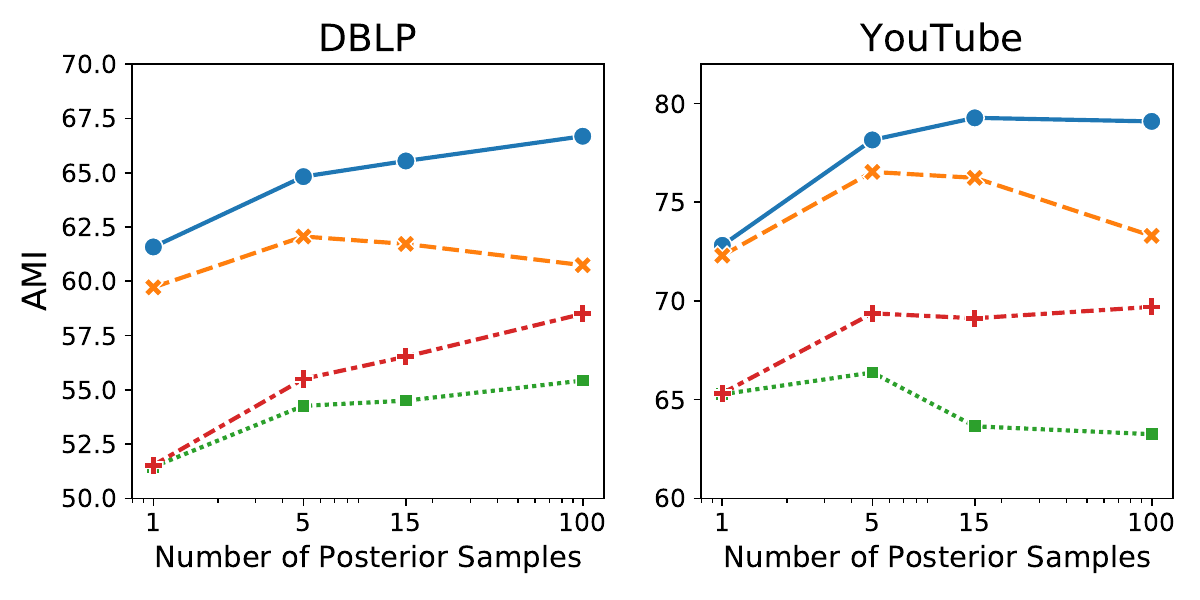}
 \caption{
    {\bf Inference time and AMI scores as a function of the number of posterior samples.} Results for two SNAP datasets with $K=2\sim4$.
   	{\it Top:} CCP is faster than NCP not only for 
   	changing $N$ or $K$, but also as a function of 
   	the number of posterior samples. 
   	{\it Bottom:} AMI scores.   	Note that when the model includes attention, the clustering quality improves with more samples; 
  	this trend is weaker or absent without attention. Each point is the mean of 4 runs.
  }  
  \label{fig:time_ami_vs_samples}
\end{figure}

\subsection{Performance on SBM and SNAP datasets}

Table \ref{tab:sbm_and_snap_results} summarizes the results of  amortized community detection on the general SBM and SNAP datasets. 
Since DMoN and LGNN require hard-coding a fixed $K$ in the neural architectures, we experimented with different $K$ values higher or equal to the maximum number of communities. The result shows that 
combining GCN with amortized clustering (e.g. GCN-CCP) learns more accurate community discovery than LGNN/GNN~\cite{chen2018supervised} in terms of AMI and ARI, and that adding attention modules to CCP gives rise to the best-performing model GCN-CCP-Attn.

\begin{figure}[h!] 
  \begin{center}
    \includegraphics[width=\linewidth]{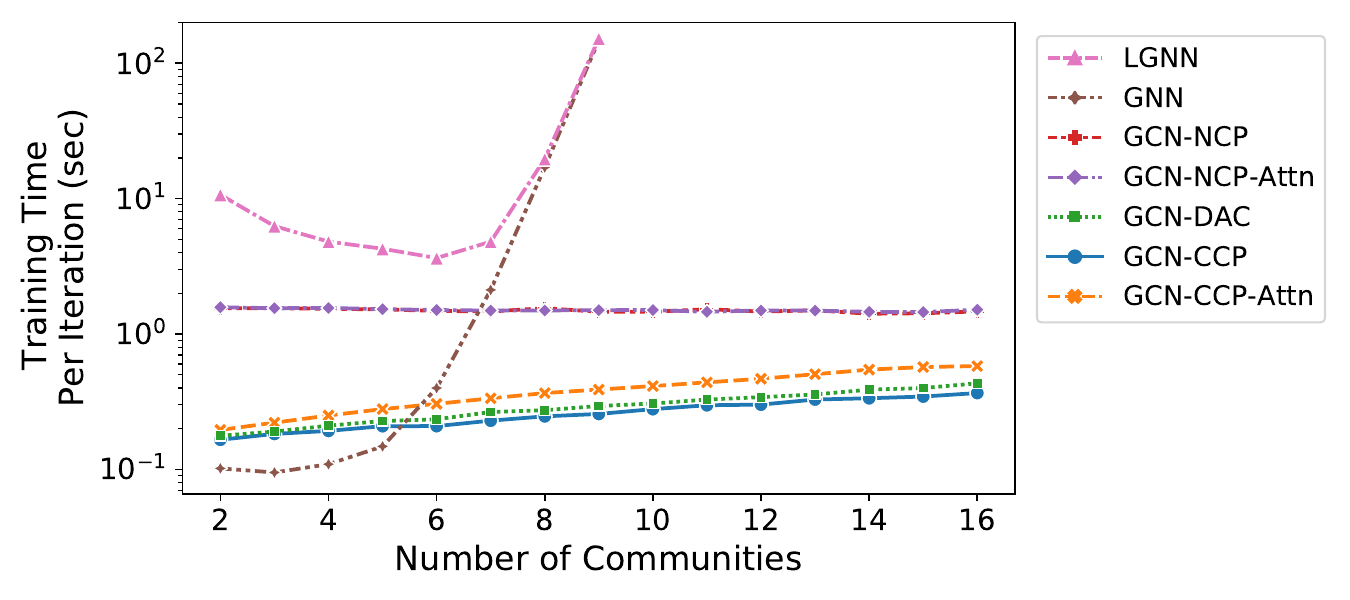}
  \end{center}
  \caption{{\bf Training times of amortized models.}  
  Per-iteration training time as a function of the number of  communities. The models are trained on the same GPU with batch size of~1 on a SBM dataset with 160 nodes and varying numbers of equal-size communities 
  ($p=0.3$, $q=0.1$). 
  }
  \label{fig:training_time}
\end{figure}

In the SBM dataset with $1\sim16$ communities, both GCN-NCP and GCN-CCP models achieved similarly good performance, and GCN-NCP outperforms by an small margin. Due to the $O(K!)$ loss function, LGNN/GNN can only train on graphs with up to 7 communities, thus cannot learn community detection at $K\geq8$.

The real-world SNAP datasets are composed of subgraphs with either fixed ($K=3$) or varying ($K=2\sim4$) numbers of communities. On all SNAP benchmarks, GCN-CCP provides consistently higher AMI and ARI than other models, and GCN-CCP-Attn further improves performance by a significantly margin. We attribute this performance gain on SNAP datasets to the expressive power of attention.

\subsection{Time Measures and MAP Estimates}
Table \ref{tab:sbm_and_snap_results} illustrates the benefits of amortization at test time, as our models run more than 10x faster than the unsupervised model (DMoN). The training time (5$\sim$10 hrs) is worth if the number of test examples is on the order of thousands. 

Figure \ref{fig:time_ami_vs_samples}, top, 
shows that CCP models scale better than NCP 
as a function of not only the number of data points, 
but also the number of posterior samples. 
Figure \ref{fig:time_ami_vs_samples}, bottom, shows that 
for  models with attention, the clustering quality improves with more samples, indicating a better fit
of the learned probabilistic model to the 
data distribution. 

Figure~\ref{fig:training_time} reports per-iteration training time. The steep time increase w.r.t. $K$ limits LGNN/GNN to graphs with less than 8 communities. The moderate growth w.r.t $K$ of 
DAC/CCP/NCP illlustrates the benefits of our architectures.






\begin{figure*}[t!]
	\begin{center}
	\fbox{
	\includegraphics[width=.465\textwidth]{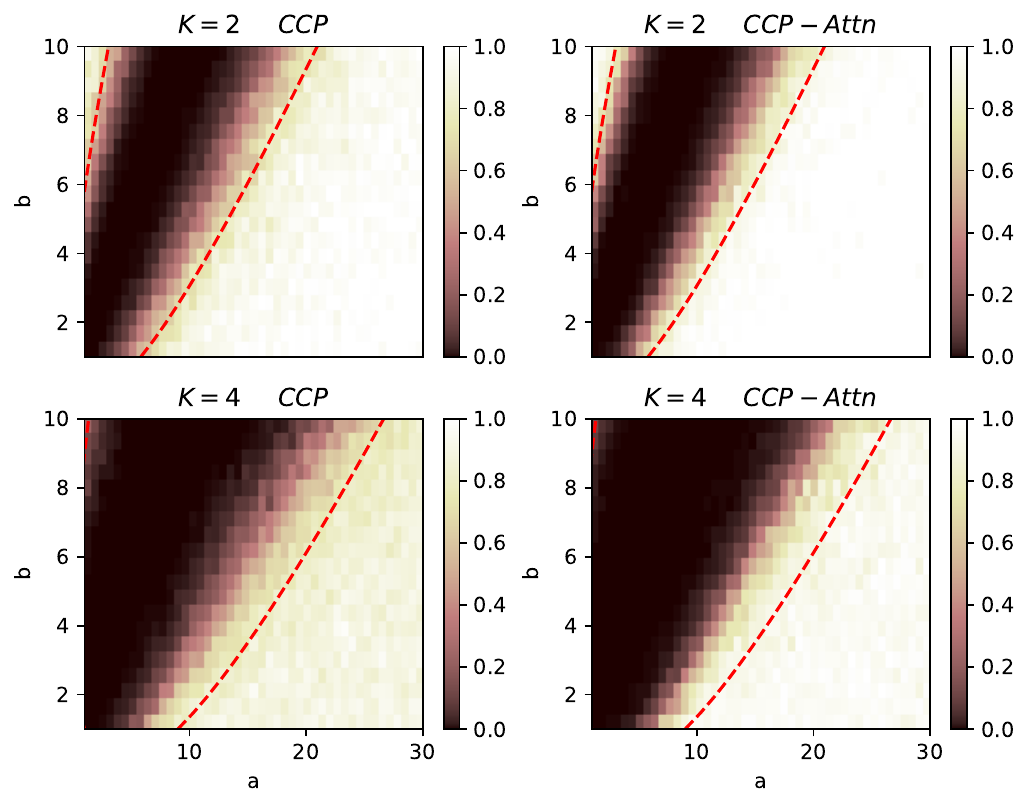}
	}
	\hfill 
	\fbox{
		\includegraphics[width=.41\textwidth,height=6.3cm]{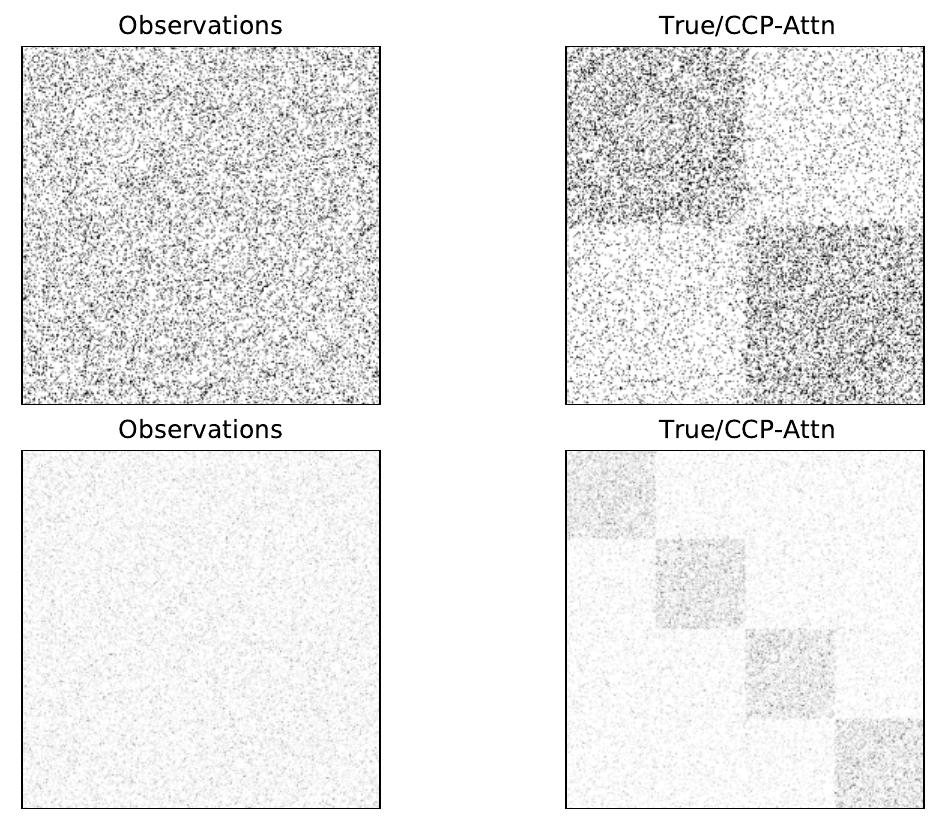}
		}
\vskip .4cm		
	\end{center}
 	\caption{{\bf Log-degree symmetric SBM.}
Each network in this family has $K$ communities  of equal size $N/K$, as illustrated in the right.  
The connection probability is $p=a \log(N)/N$ within each community and $q=b \log(N)/N$ between communities.
{\it Left:} Dashed red curves indicate the threshold
  	$|\sqrt{a} - \sqrt{b}|= \sqrt{K}$, 
  	which separates regions of possible/impossible
  	recovery with high probability using maximum-likelihood at large~$N$~\cite{abbe2015exact,mossel2014consistency,abbe2015community,abbe2015recovering}.   
  	{\it Left above:} 
  	Mean AMIs over 40 test datasets with $N=300, K=2$. 
  	The mean AMI averaged over the recoverable regions is 0.970 for
  	GCN-CCP and $0.997$ for GCN-CCP+Attn. 
  	{\it Left below:} Similar for mean of 10 datasets with $N=600, K=4$.
  	The mean AMI averaged over the recoverable regions is $0.858$ for	GCN-CCP and $0.942$ for GCN-CCP+Attn. Note that the advantage of GCN-CCP-Attn over
  	GCN-CCP is 
  	more prominent for higher $K$, 
  	and that  the thresholds are crossed  in our finite $N$ case.  
 	{\it Right:} Examples of observed 
 	adjacency matrices and 
 	successful exact recovery by GCN-CCP-Attn,
 	for $N=300,\, K=2,\, a=15,\, b=5$ ({\it above}), 
 	and $N=600,\, K=4, \, a=15,\, b=4$~({\it below}).
 	}
	\label{fig:bisection}
\end{figure*}

\subsection{Recovery Thresholds in Log-Degree  SBM}
We consider next symmetric SBMs	with $K$ communities  of equal size. 
The connection probability is 
$p=a \log(N)/N$ within 
and $q=b \log(N)/N$ between communities.
The expected degree of a node is known to be 
$O(\log N)$. 
This is an interesting regime, 
since for large $N$ 
it was shown using information-theoretic arguments that 
the maximum likelihood estimate of the $c_i$'s 
recovers exactly the community structure
with high probability for $|\sqrt{a} - \sqrt{b}|> \sqrt{K}$ and fails for $|\sqrt{a} - \sqrt{b}|< \sqrt{K}$~\cite{abbe2015exact,mossel2014consistency,abbe2015community,abbe2015recovering}. 

We trained  GCN-CCP and GCN-CCP+Attn networks
with samples generated 	with $K\in \{ 2,3,4 \}$, 
  $N \in [300,600]$ and $(a,b)$ sampled uniformly from $[1,30]\times [1,10]$. Figure~\ref{fig:bisection} shows
mean test AMI scores and examples of exact recovery.  Note in particular that improvement due to the presence of the attention modules is more prominent as~$K$ increases.

\subsection{Uncertainty quantification}
An advantage of our fully probabilistic 
model  
is the ability
to quantify the uncertainty of inferred quantities. We illustrate this in Figure~\ref{fig:sbm_uncertainty} (Appendix),
which shows the mean and 
and standard deviation of the number of clusters $K$ in a family of SBM models (see details in the Figure's caption).

\subsection{Calibration}

In order to probe 
how well calibrated the learned posterior 
distributions are, 
we calculated the expected calibration error (ECE) metric \cite{guo2017calibration} for classification
applied to the prediction of the number of 
clusters $K$. 
ECE is defined as
$
\mathrm{ECE}=\sum_{m=1}^{M} \frac{\left|B_{m}\right|}{n}\left|\operatorname{acc}\left(B_{m}\right)-\operatorname{conf}\left(B_{m}\right)\right|
$,
where $n$ is the number of samples and each $B_m$ is one of $M$ equally spaced bins, where $B_m$ is the set of indices whose confidence is inside the interval $I_{m}=\left(\frac{m-1}{M}, \frac{m}{M}\right]$.
In \Cref{tab:ece_K}, 
we compare the ECE for two of our models. 
We sampled $15$ community assignments from the posterior of each trained model and used the distribution of predictions to measure confidence.
As expected, the CCP model performs better than 
DAC, since 
its architecture encodes the correct 
probabilistic inductive biases.  
\vskip 4mm 

\begin{table}[!h] 
\centering
\caption{
{\bf Expected Calibration Error on SNAP datasets.}
Means and st. dev. of ECE from $5$ independent runs.
}
\label{tab:ece_K}
{ 
\begin{tabular}{lcc}
\toprule
& DBLP & Youtube \\
\midrule
GCN-DAC
    & \msd{0.7406}{0.1325} & \msd{0.7932}{0.0716} \\
GCN-CCP
    & \BF \msd{0.1579}{0.0295} & \BF \msd{0.1804}{0.0286} 
    \\
\bottomrule
\end{tabular} }
\end{table}

\section{Conclusion}
We have introduced a novel framework for efficiently detecting community structures in graph data, 
building on recent advances in graph neural networks and amortized clustering.
Our experiments have shown that our proposed method outperforms previous methods for community detection on many benchmarks.
Possible future directions include incorporating new combinations of attention modules, learning SBMs in the weak recovery regime~\cite{abbe}, and dealing with overlapping communities.


\bibliography{spigm_references}
\bibliographystyle{icml2024}

\newpage
\appendix
\onecolumn


\section{Experimental Details}
\subsection{General SBM Dataset}

The SBM dataset with $K = 1\sim16$ communities are created according to the generative model in Section 5.1. Communities with less than 5 nodes are removed from the resulting graphs. The train, validation and test sets contain 20000, 1000, and 1000 graphs, respectively. 

\subsection{SNAP Datasets}
The SNAP dataset \citeapp{snapnets} is distributed under the BSD license, which means that it is free for both academic and commercial use.
In each SNAP dataset of real-world graphs, we use the top $5000$ high quality communities chosen according to an average of six scoring functions in \citeapp{yang2015defining}. 
We randomly split these into $3000$ train, $500$ validation and $1500$ test communities, and extracted subgraphs composed of multiple non-overlapping communities from each split to form the train, validation and test sets. 
For the 3-community experiments, we find triplets of communities $C_1, C_2, C_3$ such that they form a connected graph, and that no nodes belong to multiple communities.
We filtered for graph size and community imbalance to ensure each pair of communities satisfies $20 < \left|C_1 \cup C_2\right| < 500$, $|C_1| < 20|C_2|$, and $|C_2| < 20|C_1|$. 
For the experiments with $2 \sim 4$ SNAP communities, we first created a community graph in which each node represents a community, and an edge exists between a pair of nodes if the two corresponding communities are not overlapping and satisfy the size/imbalance constraints above. 
We then extracted subgraphs from the each SNAP dataset by finding cliques of size 2-4 in the community graph. The dataset statistics is shown in Table \ref{tab:snap_data_stats}. For datasets with more than 1000 graphs in the test set, the test set is randomly subsetted to 1000 graphs for faster evaluation. 

\begin{table*}[h] \centering
\begin{tabular}{l ccc}
\toprule
Dataset & Train/Val/Test Networks & $|V|$ & $|E|$ \\
\midrule
DBLP ($K=3$) & $3929/28/696$ & 65 & 406 \\
DBLP ($K=2\sim4$) & $3222/166/1000$ & 100 & 542 \\
Youtube ($K=3$) & $8670/1683/1000$ & 100 & 442 \\
Youtube ($K=2\sim4$) & $22141/2687/1000$ & 122 & 559 \\
\bottomrule \end{tabular}
\vspace{-5pt}
\caption{ SNAP dataset statistics.}
\label{tab:snap_data_stats}
\end{table*}
\vspace{-5pt}

\subsection{Model Training and Inference}
All proposed models are implemented in PyTorch and trained with a batch size of 16. The number of training iterations is 5000 for the 3-community SNAP dataset, and 10000 for the SNAP dataset with $2\sim4$ communities and the general SBM dataset. The learning rate is 0.0001 for GCN-CCP, GCN-DAC, GCN-CCP-Attn, and 0.00005 for GCN-NCP and GCN-NCP-Attn.

For DMoN~\citeapp{tsitsulin2020graph}\footnote{
DMoN: \url{https://github.com/google-research/google-research/tree/master/graph_embedding/dmon}
} and LGNN/GNN~\citeapp{chen2018supervised}\footnote{
LGNN/GNN: \url{https://github.com/zhengdao-chen/GNN4CD}}, we used the official implementations
with default parameters unless otherwise noted. The DMoN model has a hidden dimension of 512 and is optimized for 1000 iterations. The LGNN/GNN models are trained for the same amount of iterations as our proposed methods on SNAP datasets, and twice amount of iterations on the SBM dataset.

All run time measurements are made on Nvidia P100 GPUs.

\section{Network Architectures}
\label{app:network_details}

\subsection{Graph Convolutional Networks (GCN)}

GCN is a class of message-passing GNN that updates the representation of each node based on local neighborhood information.

{\bf GraphSAGE.} GraphSAGE~\citeapp{hamilton2018inductive} defines a graph convolution operation that updates the features of each node by integrating the features of both the center and neighboring nodes:
\eqan 
h_i^{\ell+1} = \mathrm{ReLU} \Big( U^\ell h_i^\ell +  V^\ell \mathrm{Aggregate}_{j \in N_i} \{ h_j^\ell\} \Big)
\,,
\enan
where $h_i^\ell$ is the feature of node $i$ at layer $\ell$, $N_i$ is the neighborhood of node $i$, $U^\ell$ and $V^\ell$ are learnable weight matrices of the neural network. The neighborhood aggregation function can be a simple mean function, or more complex LSTM and pooling aggregators. GraphSAGE belongs to isotropic GCNs in which each neighbor node contributes equally to the update function. We use a 4-layer GraphSAGE GCN with the mean aggregator and batch normalization (BN) for our experiments:
\eqan 
h_i^{\ell+1} = \mathrm{ReLU} \Big( \mathrm{BN} ( U^\ell h_i^\ell +  V^\ell \mathrm{Mean}_{j \in N_i} \{ h_j^\ell\} ) \Big)
.
\enan

{\bf GatedGCN.} GatedGCN~\citeapp{bresson2018residual} is an anisotropic GCN that leverages edge gating mechanisms. Each neighboring node in the graph convolution operation may receive different weights depending on the edge gate. Residual connections are used between layers for multi-layer GatedGCN. To improve GatedGCN, \citeapp{dwivedi2020benchmarking} proposed explicitly updating edge gates across layers:
\eqan 
h_i^{\ell+1} = h_i^{\ell} + \mathrm{ReLU} \Big( \mathrm{BN} (U^\ell h_i^\ell +  \sum_{j \rightarrow i} e_{ij}^\ell \odot V^\ell h_j^\ell ) \Big)
\,,
\enan
where $h_i^0 = x_i, e_{ij}^0 = 1$, and $e_{ij}^\ell$ is the edge gate computed as follows:
\eqan 
e_{ij}^\ell &=& \frac{\sigma(\hat{e}_{ij}^\ell)}{ \sum_{j' \rightarrow i} \sigma(\hat{e}_{ij'}^\ell) + \epsilon }  
\,, \\
\hat{e}_{ij}^\ell &=& \hat{e}_{ij}^{\ell-1} + \mathrm{ReLU} \Big( \mathrm{BN} ( A^\ell h_i^{\ell-1} + B^\ell h_j^{\ell-1} + C^\ell \hat{e}_{ij}^{\ell-1} ) \Big) 
.
\enan
We used a 4-layer GatedGCN encoder with hidden dimension of 128 in each layer. 

\subsection{Attention Modules}
\label{app:attention}
Our attention modules are composed of standard Multi-Head Attention (MHA) blocks~\citeapp{vaswani2017attention} , which take 
as inputs $n$ query vectors 
${\bf q} = (q_1 \dots q_n)^\top \in \mathbb{R}^{n \times d_q}$,
$m$ key vectors ${\bf k} = (k_1 \dots k_m)^\top \in \mathbb{R}^{m \times d_k}$,
and $m$ value vectors 
${\bf v} = (v_1 \dots v_m)^\top \in \mathbb{R}^{m \times d_v}$,
and return~$n$ vectors $\mathrm{MHA}({\bf q}, {\bf k}, {\bf v}) \in \mathbb{R}^{n \times dh}$.
Using the MHA, we follow~\citeapp{lee2018set} in defining 
three attention modules for functions defined over sets:

\begin{itemize}
    \item {\bf Multihead Attention Block (MAB).}
Given  two sets 
$\x = (x_1,\dots, x_n)^\top \in \mathbb{R}^{n \times d_x}$ 
and $\y = (y_1,\dots, y_m)^\top \in \mathbb{R}^{m \times d_y}$, we define 
\eqan 
\mathrm{MAB}(\x, \y) = {\bf h} + \mathrm{FF}({\bf h}) 
\in \mathbb{R}^{n \times d}
\\
\text{ where } {\bf h} = \x W_q^{[h]} + \mathrm{MHA}(\x, \y, \y)
\in \mathbb{R}^{n \times d},
\label{eq:MAB}
\enan 
with $W_q^{[h]} = [W_q^{1} \ldots W_q^{h}] \in \mathbb{R}^{d_x \times d}$ and $\mathrm{FF}$ 
is a feed-forward layer applied to 
each row of ${\bf h}$. 
Note that in~(\ref{eq:MAB}), $\y$ is both the 
key and value set and that $\mathrm{FF}$ and $\mathrm{MHA}$ have their own trainable parameters. 

\item {\bf Pooling by Multihead Attention (PMA).}
Given a set ${\bf x}$, we create a permutation invariant summary 
into a small number of $m$ vectors 
with 
\[
\mathrm{PMA}_m(\x) = \mathrm{MAB}({\bf e}, \x) \in \mathbb{R}^{m \times d}
\]
where ${\bf e} = (e_1, \dots, e_m)  \in \mathbb{R}^{m \times d}$ 
are trainable parameters. This is a weighted pooling of the items in~$\x$, where an attention mechanism determines the weights.

\item {\bf Induced Self-Attention Block (ISAB).}
The time complexity of expressing self-interactions in a set via $\mathrm{MAB}(\x, \x)$
scales as $O(n^2)$.
To reduce this cost we 
approximate the full pairwise comparison 
via a smaller trainable set of inducing points 
 ${\bf s} = (s_1, \dots, s_m)$,
\eqan 
\mathrm{ISAB}(\x) 
&=& \mathrm{MAB}(\x, \mathrm{MAB}({\bf s}, \x) ) 
\in \mathbb{R}^{n \times d}
\nn 
\enan 
Thus we indirectly compare 
the pairs in $\x$ using~${\bf s}$ as a bottleneck, with  time complexity $O(nm)$. 
Since $\mathrm{PMA}_m({\x})$ is invariant 
under permutations of ${\bf x}$'s  rows, 
 ISAB is permutation-equivariant. 
Higher-order interactions 
are  obtained by stacking multiple ISAB layers.
\end{itemize}


\subsection{CCP Architecture}
\label{app:ccp-arch}

The CCP model is implemented as described in \citeapp{pakman2020}. The full CCP architecture, including the prior, likelihood and posterior components, is illustrated in Figure \ref{fig:ccp_model_with_posterior}. The posterior network is only used in training, thus not displayed in the 
diagram of Figure 2 in the main text. 

\subsubsection{Encodings}
In order to parametrize the  prior, likelihood and posterior of the CCP model, it is convenient to define first 
some symmetric encodings 
for different subsets of the dataset $\x$ at iteration $k$.

In the main text we used $x_a$  to refer to the 
first element in cluster $\s_k$,
and $x_{q_i}$ to indicate the additional points 
available to join $\s_k$. 
Here we use instead $x_{d_k}, x_{a_i}$ respectively, 
in order to be consistent with the notation of~\citeapp{pakman2020}.
The notation $\x_k$ indicates that 
the dataset is split into three groups, $\x_k = (\x_{\bfa},x_{d_k},\x_{\s})$, where

\begin{center}
\begin{tabular}{ll}
    $\x_{\bfa}$= $ (x_{a_1} \ldots x_{a_{m_k}})$ 
    & $m_k$ available points for  cluster k 
    \\
    $x_{d_k}$&  First data point  in cluster k
    \\
    $\x_{\s}=(\x_{\s_1} \ldots \x_{\s_{k-1}} )$&  
    Points already assigned to clusters. 
\end{tabular}
\end{center}
Let us also define
\eqan 
 \bar{\bf u}_{\bf a} = (\bar{u}_1 \ldots \bar{u}_{m_k}) = u(x_1) \ldots  u(x_{m_k}) 
\label{eq:ccp_u}
\enan

The  encodings we need are:

\everymath{\displaystyle}

\begin{table}[h!]
\begin{center}
\begin{equation}
\begin{array}{|cl|l|}
\hline 
& \text{Definition}  & \text{Encoded Points} 
\\ 
\hline 
D_k & = x_{d_k}
&
x_{d_k}, \text{ the first point in cluster k}
\\
U_k &=  \mathrm{mean}(\bar{\bf u}_{\bf a})
&  \x_{\bfa}, \text{all the $m_k$ points available to join } x_{d_k} 
\\
U_k^{in} &= \mathrm{mean} \Big( \bar{u}_{a_i},  i \in (1 \ldots m_k), b_i = 1 \Big)
&
\text{Points from $\x_{\bfa}$ that join cluster $k$.}
\\
U_k^{out} &= \mathrm{mean} \Big( \bar{u}_{a_i}, i \in (1 \ldots m_k), b_i = 0 \Big)
& 
\text{Points from $\x_{\bfa}$ that do not join
cluster $k$}
\\
G_k &= \sum_{j=1}^{k-1} g \Big( \mathrm{mean} \Big( h({x}_{i}), i \in \s_j \Big)
&
\text{All the clusters } \s_{1:k-1}.
\\
\hline 
\end{array}
\label{table:ccp_encoding_mlp}
\end{equation}
\end{center}
\end{table}

\subsubsection{Prior and Likelihood}


Having generated $k-1$ clusters $\s_{1:k-1}$,
the  elements of $\s_k$ are generated in a process with latent variables $d_k,\z_k$ 
and joint distribution. 
\eqan 
\quad & p_{\theta}(\s_k,\z_k,d_k|\s_{1:k-1},\x)=
p_{\theta}(\bb_k|\z_k, \x_k) p_{\theta}(\z_k|\x_k)p(d_k|\s_{1:k-1}) \,,
\enan 
where 
\eqan 
p_{\theta}(\bb_k|\z_k, \x_k)= \prod_{i=1}^{m_k}
\varphi(b_{i}|\z_k, U, G, x_{d_k}, x_{a_i} )  
\,.
\enan

The priors and likelihood are
\eqan
p(d_k|\s_{1:k-1}) &=& \left\{
\begin{tabular}{cc}
    $1/ |I_k|$ & \textrm{for} $d_k \in I_k$ ,
    \\
    0 &  \textrm{for} $d_k \notin I_k$  ,
\end{tabular}
\right.
\\
p_{\theta}(\z_k|\x_k) &=& 
{\cal N}(\z_k|\mu(\x_k) , \sigma(\x_k))
\label{eq:p1}
\\
\varphi(b_{i}|\z_k, U, G, x_{d_k}, x_{a_i} )  
&=& \textrm{sigmoid}[ \rho_i(\z_k, \x_k)] 
\label{eq:p2}
\enan
where $I_k$ is the set of indices available
to become the first element of $\s_k$, 
and we have defined 
\eqan 
\mu(\x_k)&=&\mu(D_k, U_k, G_k)
\\
\sigma(\x_k)&=&\sigma(D_k, U_k, G_k), 
\\
 \rho_i(\z_k, \x_k) &=&  \rho(\z_k, x_{a_i}, D_k, U_k, G_k)
 \qquad i=1 \ldots m_k
\enan 
where $\mu, \sigma,\rho$
are represented with MLPs. 
Note that in all the cases the functions 
depend on encodings in (\ref{table:ccp_encoding_mlp}) that are consistent with the permutation symmetries dictated by the conditioning information.

\subsubsection{Posterior}

To learn the prior and likelihood functions, we introduce 
\eqan
q_{\phi}(\z_k,d_k|\s_{1:k}, \x) =q_{\phi}(\z_k|\bb_k, d_k, \x_k)
q_{\phi}(d_k|\s_{1:k}, \x)
\enan
to approximate the intractable posterior. This allows us to train CCP as a conditional variational autoencoder (VAE) \citeapp{sohn2015learning}.

For the first factor we assume a form
\eqan 
q_{\phi}(\z_k|\bb_k, d_k, \x_k) &=&
{\cal N}(\z_k|\mu_q(D_k, A_k^{in}, A_k^{out},G_k), 
\sigma_q(D_k, A_k^{in}, A_k^{out},G_k))
\label{eq:q}
\enan 
where $\mu_q,\sigma_q$ are MLPs.
For the second factor we assume
\eqan
q(d_k|\s_{1:k}) &=& \left\{
\begin{tabular}{cc}
    $1/N_k$ & \textrm{for} $d_k \in \s_k$,
    \\
    0 &  \textrm{for} $d_k \notin \s_k$.
\end{tabular}
\right.
\enan 
This approximation is very good in
cases of well separated clusters.
Since $q(d_k|\s_{1:k})$ has no parameters,
this avoids the problem of backpropagation through discrete variables. 

\subsubsection{ELBO}
The ELBO that we want to maximize is given by 
\eqan
&& 
\mathbb{E}_{p(\x,\s_{1:K})}  \log p_{\theta}(\s_{1:K}|\x) 
\\
&& = 
\mathbb{E}_{p(\x,\s_{1:K})}
\sum_{k=1}^K 
\log\left[\sum_{d_k=1}^{N_k}  \! \! \int \! \! d\z_k  p_{\theta}(\s_k,\z_k,d_k|\s_{1:k-1},\x)
\right]
\\
&& 
\geq 
\mathbb{E}_{p(\x,\s_{1:K})} 
\sum_{k=1}^K 
\mathbb{E}_{q_{\phi}(\z_k,d_k|\s_{1:k},\x)}
\log \left[
\frac{p_{\theta}(\s_k,\z_k,d_k|\s_{1:k-1},\x)}
  {q_{\phi}(\z_k,d_k|\s_{1:k},\x)}
 \right]
\\
&& 
=
\mathbb{E}_{p(\x,\s_{1:K})} 
\sum_{k=1}^K 
\mathbb{E}_{q_{\phi}(\z_k,d_k|\s_{1:k},\x)}
\log \left[
\frac{
p_{\theta}(\bb_k|\z_k, \x_k) p_{\theta}(\z_k|\x_k)p(d_k|\s_{1:k-1})
}  
{
q_{\phi}(\z_k|\bb_k, d_k, \x_k)q_{\phi}(d_k|\s_{1:k}, \x)
}
 \right]
\enan

\begin{figure}[t!]
\centering \includegraphics[width=.98\textwidth]{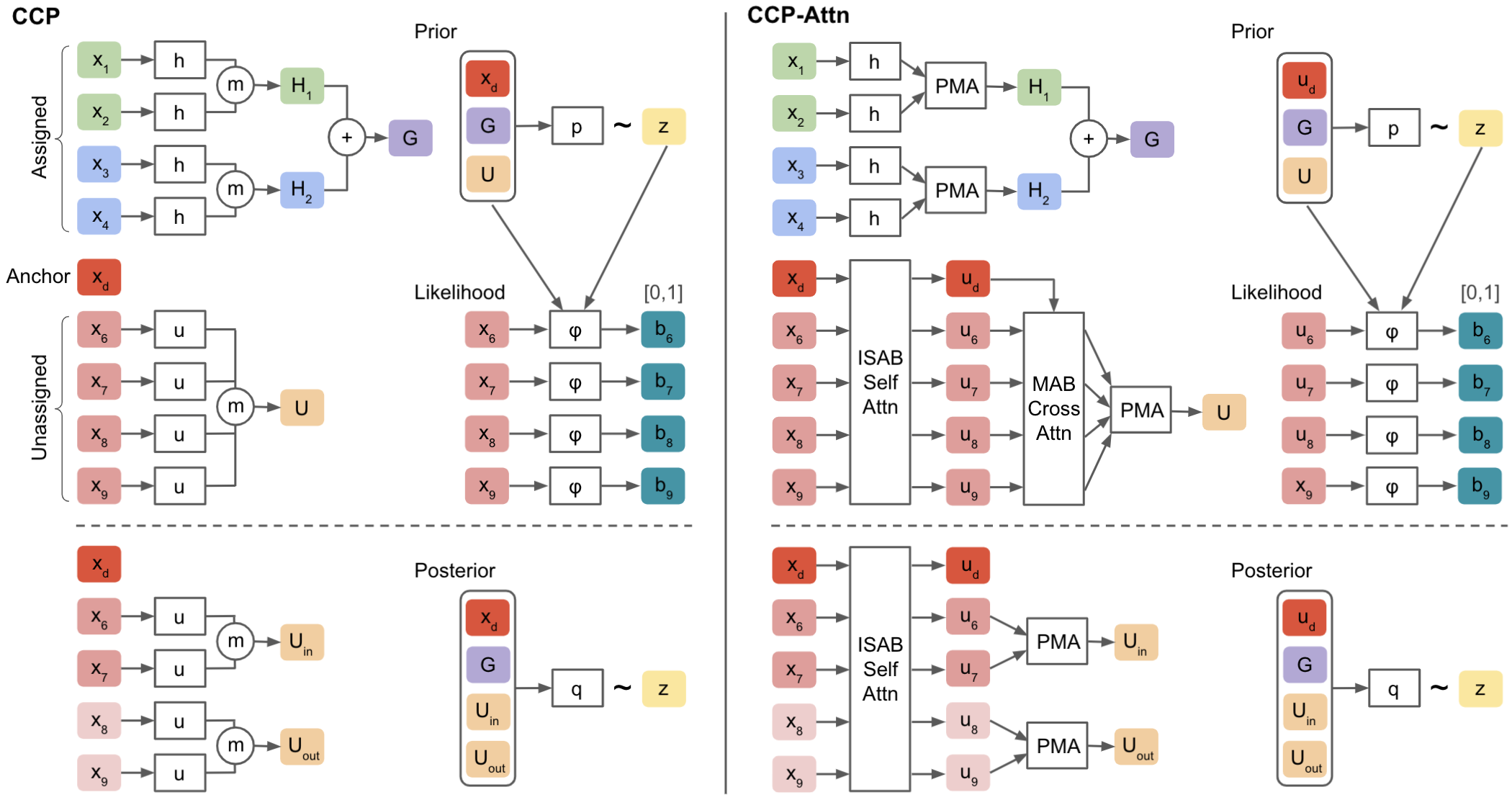}
 \caption{
  	{\bf Full architecture of CCP and CCP-Attn.} The diagram illustrates the conditional prior, likelihood and posterior components of CCP and CCP-Attn. The mean aggregations \textcircled{\scriptsize m} used by CCP (see equation 
  	are replaced in CCP-Attn by Set Transformer attention modules from~\citeapp{lee2018set}.
  }  
  \label{fig:ccp_model_with_posterior}
\end{figure}

\subsection{CCP-Attn Architecture}
\label{app:ccp-attn-arch}

In CCP-Attn, we replace equation (\ref{eq:ccp_u}) with a ISAB self attention layer among all available points:

\eqan 
(\bar{u}_{d_k}, \bar{u}_1 \ldots \bar{u}_{m_k})
 &=& \mathrm{ISAB} [ u(x_{d_k}),u(x_1) \ldots  u(x_{m_k}) ]  
 \\
 \bar{\bf u}_{\bf a} &=& (\bar{u}_1 \ldots \bar{u}_{m_k})
\nn 
\enan 

and replace the encodings in (\ref{table:ccp_encoding_mlp}) with attention-based aggregations:

\everymath{\displaystyle}

\begin{table}[h!]
\begin{center}
\begin{equation}
\begin{array}{|rl|l|}
\hline 
& \text{Definition}  & \text{Encoded Points} 
\\ 
\hline 
D_k & = \bar{u}_{d_k}
&
x_{d_k}, \text{ the first point in cluster k}
\\
U_k &=  \mathrm{PMA}(\mathrm{MAB}(\bar{\bf u}_{\bf a},\bar{u}_d)) 
&  \x_{\bfa}, \text{all the $m_k$ points available to join } x_{d_k} 
\\
A_k^{in} &= \mathrm{PMA} \Big( \bar{u}_{a_i},  i \in (1 \ldots m_k), b_i = 1 \Big)
&
\text{Points from $\x_{\bfa}$ that join cluster $k$.}
\\
A_k^{out} &= \mathrm{PMA} \Big( \bar{u}_{a_i}, i \in (1 \ldots m_k), b_i = 0 \Big)
& 
\text{Points from $\x_{\bfa}$ that do not join
cluster $k$}
\\
G_k &= \sum_{j=1}^{k-1} g \Big( \mathrm{PMA} \Big( h({x}_{i}), i \in \s_j \Big)
&
\text{All the clusters } \s_{1:k-1}.
\\
\hline 
\end{array}
\label{table:encoding_attn}
\end{equation}
\end{center}
\end{table}

Together, these changes give rise to the CCP-Attn architecture illustrated in Figure \ref{fig:ccp_model_with_posterior}.

\subsubsection{Neural Networks in CCP and CCP-Attn}
\label{app:ccp-network-layers}

The $h$, $g$ and $u$ functions are parameterized by 3-layer MLPs. The $p$, $q$ and $\varphi$ functions are MLPs with 5, 5, and 4 layers, respectively. All MLPs have hidden-layer dimensions of 128 and parametric ReLU (PReLU) layers in between linear layers. Each vector in the equations above has a dimension of 128. In CCP-Attn, the ISAB, MAB, and PMA attention modules contain 32 inducing points, 4 attention heads,  and hidden-layer dimensions of 128. 

\subsection{NCP and NCP-Attn Architecture}
\label{app:ncp-arch}

The NCP architecture is described in \citeapp{pakman2020}. In NCP, each cluster $k$ is encoded by permutation-invariant aggregation of data points assigned to that cluster
\eqan
H_k= \sum_{i : c_{i}=k} h(x_i)  
\quad h:\mathbb{R}^{d_x} \rightarrow \mathbb{R}^{d_h}.
\label{Hk}
\enan 

The global representation of the current clustering configuration is given by 
\eqan 
G= \sum_{k =1}^K g(H_k) ,
\quad 
g:\mathbb{R}^{d_h} \rightarrow \mathbb{R}^{d_g}.
\label{G_def}
\enan

Given $n-1$ assigned points $x_{1:n-1}$ and their cluster labels $c_{1:n-1}$, we want to find the cluster label for the next point $x_n$. At this point, the unassigned points $x_{n+1:N}$ are represented by
\eqan 
U = \sum_{i=n+1}^{N}  u(x_i) \,,
	\qquad 
u:\mathbb{R}^{d_x} \rightarrow \mathbb{R}^{d_u}.
\label{Q}
\enan 

The probability of the next point $x_n$ joining cluster $k$ is modeled by the variable-input softmax function 
\eqan 
q_{\theta}(c_n=k|c_{1:n-1}, \mathbf{ x}) = \frac{ e^{f(G_k,U)} }
{  \sum_{k'=1}^{K+1} e^{f(G_{k'},U)}       }.
\label{ncp}
\enan

The $h$, $u$, $g$, $f$ functions in NCP are MLPs with 2, 2, 5, and 5 layers, respectively. The MLPs have hidden-layer dimensions of 128 and parametric ReLU (PReLU) layers in between linear layers. In NCP-Attn, the $h$ and $u$ functions are replaced by ISAB attention across all points. The ISAB attention module contains 32 inducing points, 4 attention heads,  and hidden-layer dimensions of 128.

\subsection{DAC Architecture}
\label{app:dac-arch}

The DAC model is composed of the same neural network backbone as CCP-Attn, and the binary cross entropy loss function of the Anchored Filtering method in \citeapp{DAC}.

\section{Additional Experiments}
\label{app:additional-experiments}

\begin{figure}[h!]
	\begin{center}
\fbox{
\includegraphics[width=.46\textwidth]{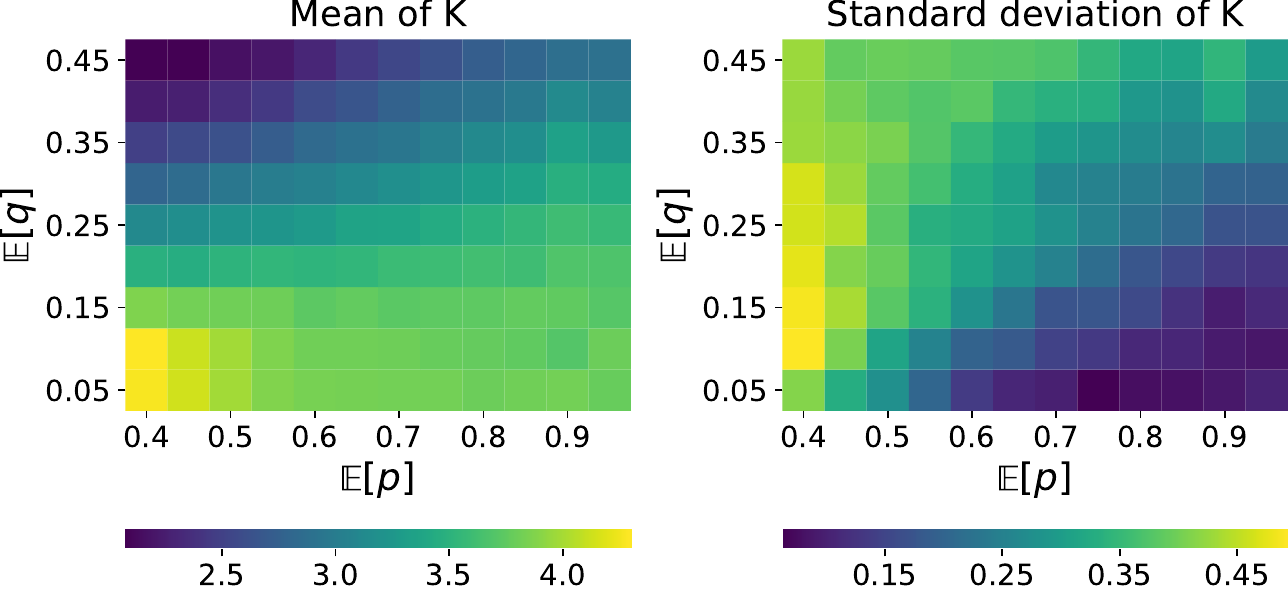}
}
	\end{center}
 	\caption{{\bf Quantifying uncertainty of inference on General SBM.} Mean ({\it left}) and standard deviation ({\it right}) of the inferred number of clusters $K$ across 500 posterior samples. The model is trained on SBM data generated from $N \sim \textrm{Unif} [50,300]$, $c_1 \ldots c_N \sim  \textrm{CRP}(0.7) $,   $p = \phi_{k_1,k_2} \sim$ \textrm{Beta}(6, 3) for $k_1 = k_2$, and $q = \phi_{k_1,k_2} \sim$ \textrm{Beta}(1, 5) for $k_1 \neq k_2$. The inference is run on SBM graphs ($N=200$, $K=4$, equal partition) with varying connection probabilities $p = \phi_{k_1 = k_2}$ and $q = \phi_{k_1 \neq k_2}$ generated from $\phi_{k_1, k_2} \sim \textrm{Beta}(\alpha, \beta)$ with $\alpha + \beta = 10$. The results are averaged over 100 test examples.
 	As expected, the standard-deviation is takes minimum values for large mean $p$ and small mean $q$.
  	} 
	\label{fig:sbm_uncertainty}
\end{figure}

Figure \ref{fig:sbm_uncertainty} illustrates how our Bayesian approach allows quantifying the uncertainty in the number of
clusters for a range of parameters in this model.

\bibliographyapp{spigm_references}
\bibliographystyleapp{icml2024}

\end{document}